\newif\ifarxiv
\arxivtrue      

\ifarxiv
  \documentclass{article}

\PassOptionsToPackage{numbers,compress}{natbib}
\usepackage[preprint]{neurips_2026}

\usepackage[utf8]{inputenc} 
\usepackage[T1]{fontenc}    
\usepackage{hyperref}       
\usepackage{url}            
\usepackage{booktabs}       
\usepackage{amsfonts}       
\usepackage{nicefrac}       
\usepackage{microtype}      
\usepackage{xcolor}         

\usepackage{graphicx,subcaption}
\usepackage{enumitem}
\usepackage{algorithm}
\usepackage{algorithmic}
\usepackage{booktabs}  
\usepackage{ragged2e}
\usepackage{wrapfig}
\usepackage{arydshln}
\usepackage{setspace}
\usepackage{adjustbox} 
\usepackage{framed}
\usepackage{tikz}
\usepackage{tabularx}
\usepackage{amssymb}
\usepackage{mathtools}
\usepackage{amsthm}
\usepackage{xcolor}
\usetikzlibrary{matrix,calc}
\usepackage{multirow}
\usepackage{arydshln}
\usepackage{subcaption}
\usepackage{rotating} 
\usepackage{booktabs}
\usepackage{makecell}
\usepackage{xcolor}
\usepackage{graphicx}
\usepackage{caption}
\usepackage{pdflscape}
\usepackage{longtable}
\usepackage{array}
\usepackage{booktabs}
\usepackage{ragged2e}
\usepackage{placeins}
\newcolumntype{P}[1]{>{\RaggedRight\arraybackslash}p{#1}}

\usepackage[table]{xcolor}
\usepackage{soul}
\usetikzlibrary{tikzmark,calc,arrows.meta}
\usetikzlibrary{positioning,calc,fit,backgrounds}
\definecolor{Ucolor}{RGB}{232,242,255}  
\definecolor{Rcolor}{RGB}{210,230,215}
\definecolor{Scolor}{RGB}{255,238,232}
\definecolor{pidHLblue}{RGB}{198,220,255}
\definecolor{pidHLgreen}{RGB}{206,242,206}
\definecolor{pidHLpink}{RGB}{255,210,225}
\definecolor{pidHLorange}{RGB}{255,225,190}
\definecolor{boxgray}{RGB}{240,240,240}

\newcommand{\kk}[1]{\textcolor{blue}{kk: #1}}

\theoremstyle{remark}

\usepackage[most]{tcolorbox}
\usepackage{calc}

\definecolor{titlebg}{HTML}{E6E5EF}    
\definecolor{linkblue}{HTML}{116E8A}

\makeatletter
\newcommand{\fairauthors}[1]{\gdef\@fairauthors{#1}}
\newcommand{\fairaffiliations}[1]{\gdef\@fairaffiliations{#1}}
\newcommand{\fairabstract}[1]{\gdef\@fairabstract{#1}}
\newcommand{\faircorrespondence}[1]{\gdef\@faircorrespondence{#1}}
\gdef\@fairauthors{}\gdef\@fairaffiliations{}
\gdef\@fairabstract{}\gdef\@faircorrespondence{}

\newcommand{\synibmaketitle}{%
  \thispagestyle{empty}%
  \begingroup
  \tcbset{enhanced,frame hidden,
          left=0.7cm,right=0.7cm,top=0.32cm,bottom=0.28cm,
          arc=5pt,colback=titlebg,
          before skip=0pt,after skip=0.3cm,
          grow to left by=1.5pt,grow to right by=1.5pt}
  \begin{tcolorbox}
    \setlength{\parindent}{0pt}%
    \justifying
    {\Large\bfseries \@title\par}
    \vskip 0.16cm
    {\small\bfseries \@fairauthors\par}
    \vskip 0.06cm
    {\small \@fairaffiliations\par}
    \vskip 0.16cm
    {\normalsize \@fairabstract\par}
    \vskip 0.16cm
    \noindent
    \begin{minipage}[c]{0.48\linewidth}
      \raggedright
      \ifx\@faircorrespondence\@empty\else
        {\small\textbf{Correspondence:} \@faircorrespondence}%
      \fi
    \end{minipage}%
    \hfill
    \begin{minipage}[c]{0.50\linewidth}
     \raggedleft
     \resizebox{0.85\linewidth}{!}{%
     \includegraphics[height=2.95cm]{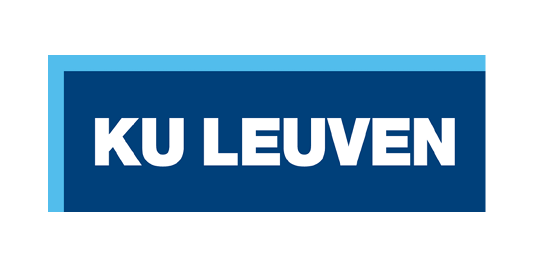}%
     \hspace{0.85cm}%
     \includegraphics[height=2.95cm]{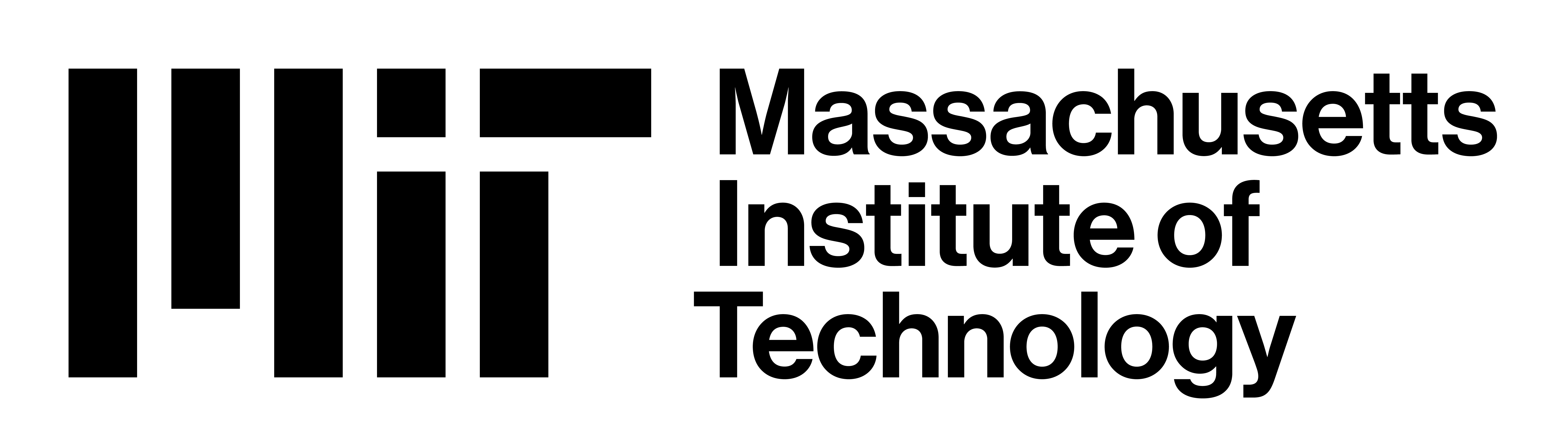}%
     \hspace{0.85cm}%
     \includegraphics[height=2.95cm]{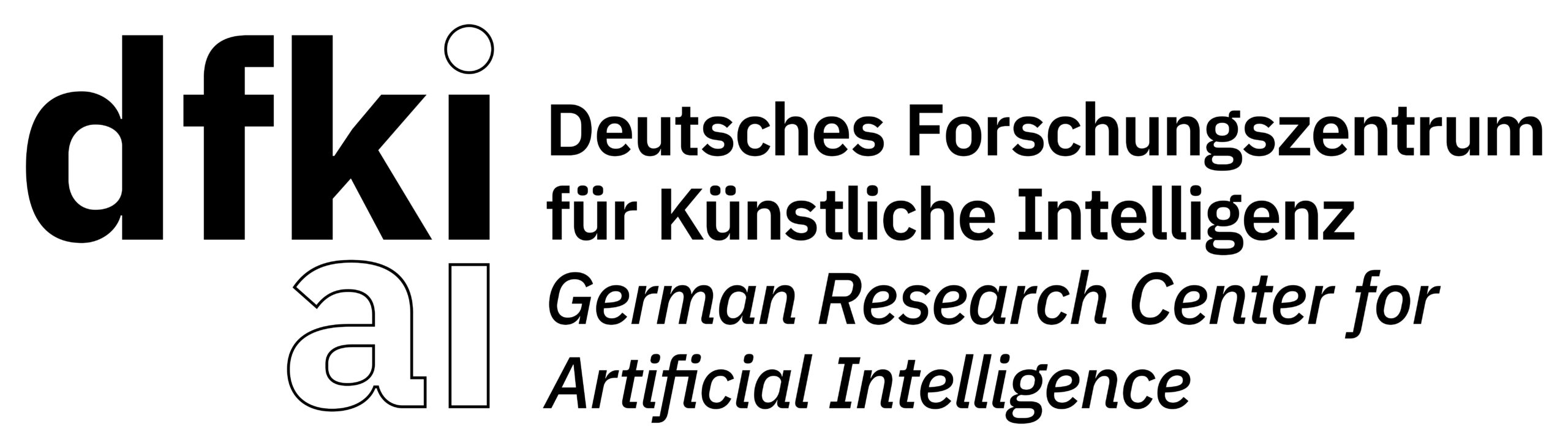}%
     }%
    \end{minipage}
  \end{tcolorbox}
  \endgroup
}
\makeatother

\title{Information-theoretic Multimodal Representation Learning for Electrocardiogram Signals}
\author{}

\fairauthors{%
  Phu X. Nguyen$^{1}$,
  Konstantinos Kontras$^{1,3}$,
  Wei Dai$^{3}$,
  Huy Phan$^{4}$,
  Christos Chatzichristos$^{1}$,
  Paul Pu Liang$^{3}$,
  Bert Vandenberk$^{1,2}$,
  Maarten De Vos$^{1}$%
}
\fairaffiliations{%
  $^{1}$KU Leuven,\quad
  $^{2}$University Hospitals
Leuven,\quad
  $^{3}$MIT, \quad
  $^{4}$DFKI%
}
\faircorrespondence{\href{mailto:phu.nguyen@kuleuven.be}{\texttt{phu.nguyen@kuleuven.be}}}

\fairabstract{
The electrocardiogram (ECG) is a widely used non-invasive measurement of cardiac activity and plays a central role in clinical diagnosis.  Recent multimodal approaches align ECG signals with clinical reports to incorporate diagnostic information. However, clinical reports primarily focus on high-level descriptions and may not fully capture the underlying physiological structure of ECG signals. This can lead to tension when models prioritize cross-modal alignment at the expense of retaining signal-specific information, particularly when modeling ECG signals across multiple levels of abstraction, from coarse diagnostic categories (e.g., rhythm or conduction abnormalities) to fine-grained waveform morphology. We aim to develop a more effective pretraining framework for ECG representation learning by integrating clinical knowledge while preserving intrinsic signal structure. We formulate this objective from an information-theoretic perspective and derive a tractable approximation that combines signal reconstruction with cross-modal contrastive alignment. Based on this principle, we propose \textbf{MERIT} (Multimodal ECG Representation via Information Theory), a dual-branch pretraining framework that integrates masked ECG modeling for structural preservation and ECG–text alignment for semantic integration. Experiments on PTB-XL and additional benchmarks demonstrate consistent improvements over prior methods, including over $3\%$ F1 on PTB-XL All and $5\%$ F1 improvement on SubClass classification. In zero-shot evaluation, the model improves performance by up to $+2.66\%$ AUC and $+2.11\%$ F1 on PTB-XL SubClass. These gains further extend to robust performance in multiple distribution-shift evaluations. In addition, using the learned ECG representations for downstream ECG-conditioned clinical text generation with LLMs improves text quality as measured with several metrics, including ROUGE (from $24.35\%$ to $25.24\%$) and METEOR (from $32.37\%$ to $33.35\%$). Together, these results convincingly demonstrate improved ECG representation quality, particularly useful for fine-grained clinical tasks.  
}

\begin{document}

\synibmaketitle

\section{Introduction} 
Electrocardiogram (ECG) analysis has achieved significant progress with deep learning models for clinical diagnosis \cite{muhammad2023, khiem2023, shengnan2023, allam2023, yang2023, wei2024, hany2024, tang2025, xiaoyu2021}. Recent self-supervised learning (SSL) approaches further improve ECG representation learning by leveraging large amounts of unlabeled data \cite{gopal20213kgcontrastivelearning12lead, clocs2021, crystal2022, pritam2022, Soltanieh2022, MaeFE2022, zhang2024, STMEM2024, HeartLang2025, ecg-jepa-kim2024, ecg-jepa-kuba2024, ecg-soup, ecg-fm2025}. While these methods effectively capture structural patterns in ECG signals, they lack explicit semantic grounding in clinical knowledge. Multimodal approaches address this limitation by aligning ECG signals with clinical reports, enabling semantic grounding and zero-shot inference \cite{junli2023, merl2024, esi2024, dbeta2025, melp2025}. However, the effectiveness of this alignment is limited by the mismatch between ECG signals and clinical text. ECG signals contain complex physiological patterns, including spatial correlations across leads and temporal dynamics in waveform morphology and rhythm. Such intrinsic structural properties of ECG signals are shared across both healthy and pathological cases and are rarely explicitly described in clinical reports. For example, ECG segments from the same lead or from related lead groups (e.g., precordial or limb leads) exhibit strong correlations, and signals aligned at the same time point show consistent temporal correspondence across leads  \cite{STMEM2024, ecg-soup}. In contrast, clinical reports typically describe diagnostic findings such as atrial fibrillation, left ventricular hypertrophy, or ST depression, focusing on diagnostic information rather than the underlying signal characteristics. As a result, aligning ECG representations with text may emphasize clinical information while overlooking signal-specific patterns that are not captured in the reports. A further challenge arises from the multi-level structure of ECG labels across different levels of abstraction. At a coarse level, labels such as rhythm and superclass capture high-level diagnostic concepts. At a finer level, sub-diagnostic and morphological labels (e.g., subclass and form) reflect subtle waveform variations. Moreover, comprehensive label spaces such as PTB-XL All include a wide range of detailed heart conditions, requiring models to capture both global and fine-grained ECG characteristics. This multi-level nature makes ECG representation learning particularly challenging, as effective models must perform consistently across heterogeneous label granularities. In particular, alignment-focused approaches may favor high-level diagnostic semantics while failing to retain fine-grained physiological details that are critical for more detailed classification tasks.

These limitations suggest a fundamental challenge: \textit{how can we learn representations that both preserve intrinsic ECG structure and align with clinical semantics?} From an information-theoretic perspective, an effective ECG representation should retain physiological information about the original signal while also capturing information shared with clinical report \cite{tishby2000informationbottleneckmethod, tishby2015deeplearninginformationbottleneck, hjelm2019learningdeeprepresentationsmutual, lele2025}. To address this, we propose a multimodal pretraining framework that jointly learns from ECG signals and clinical text while preserving signal structure. The goal is to learn representations that retain information from the ECG signal while incorporating clinically meaningful semantics. Empirical results show that balancing these two aspects improves performance, particularly in fine-grained classification and zero-shot settings.

Our contributions are as follows:
\begin{itemize}[leftmargin=*, itemsep=0.1em, topsep=0pt, parsep=0pt, partopsep=0pt]
    \item Identification of a fundamental trade-off in multimodal ECG representation learning: cross-modal alignment with clinical text can suppress ECG-specific structural information, while signal-only objectives preserve structure but lack semantic grounding. This trade-off is formalized from an information-theoretic perspective, leading to a tractable objective that unifies signal reconstruction and cross-modal alignment.

    \item A multimodal pretraining framework that jointly learns from ECG signals and clinical text, enabling representations to retain physiological structure while incorporating clinical semantics.

    \item Empirical evidence demonstrating consistent improvements across evaluation settings. The proposed approach achieves over $3\%$ F1 improvement on PTB-XL All and over $5\%$ F1 improvement on Sub-class classification in linear probing. In zero-shot evaluation, it further improves performance by up to $+2.66\%$ AUC and $+2.11\%$ F1 on Sub-class, while maintaining competitive performance on PTB-XL All. Additional experiments under distribution shift further demonstrate competitive cross-domain generalization.

    \item Extension of the proposed framework to ECG-conditioned clinical text generation with LLMs, showing that the learned ECG representations provide effective conditioning signals for downstream language generation tasks, improving ROUGE from $24.35\%$ to $25.24\%$ and METEOR from $32.37\%$ to $33.35\%$.
\end{itemize}

\section{Related Work}
We review prior work on ECG representation learning from two directions: unimodal SSL and multimodal ECG-text learning.

\subsection{Unimodal ECG Self-supervised Learning}
Unimodal SSL has been widely adopted for ECG representation learning through contrastive and generative objectives. Contrastive approaches \cite{ting2020, oord2019representationlearningcontrastivepredictive, temesgen2022} learn representations by enforcing consistency across augmented views, but may focus on invariances induced by augmentations rather than preserving fine-grained signal details \cite{STMEM2024, merl2024, xiang2024, ecg-soup}. Generative approaches, such as masked modeling \cite{STMEM2024, ecg-soup, ecg-fm2025}, reconstruct the input signal to capture temporal and spatial dependencies, leading to strong structural representations and improved downstream performance. However, since these methods rely solely on ECG signals, they lack explicit semantic grounding and are limited in their ability to connect learned representations with clinical concepts.

\subsection{ECG-text Multimodal Representation Learning}
Multimodal approaches extend ECG representation learning by incorporating clinical reports that describe diagnostic findings, such as arrhythmias, conduction abnormalities, and other clinically relevant conditions. Early works such as METS \cite{junli2023} and MERL \cite{merl2024} align ECG signals with reports via contrastive learning, thereby improving performance across different clinical tasks and enabling zero-shot transfer. Subsequent methods enhance this paradigm with additional modeling strategies, including retrieval-augmented generation ESI \cite{esi2024} and hybrid generative–contrastive objectives D-BETA \cite{dbeta2025} and MELP \cite{melp2025}. While effective in capturing diagnostic information, these approaches primarily rely on cross-modal alignment, which does not explicitly constrain representations to capture intrinsic signal structure. Generative approaches based on large language models (LLMs) further extend this paradigm by fusing ECG and text representations and optimizing for report generation or reasoning \cite{ecg-byte2024, ecg-chat2025, lai2025medr1, qoq2025}. Unlike multimodal representation learning approaches that explicitly optimize shared ECG-text embeddings for transfer and retrieval, these methods primarily focus on generation quality. As a result, they may be less suitable for tasks that require transferable and information-rich ECG representations, such as zero-shot classification or cross-task generalization.

\section{Method}

\paragraph{Problem setup.}
We consider a paired ECG--text dataset $\mathcal{D} = \{(\mathbf{X}_i, \mathbf{Y}_i)\}_{i=1}^{N}$, where $\mathbf{X}_i \in \mathbb{R}^{C \times T}$ is a 12-lead ECG signal with $C = 12$ leads and $T$ time steps and $\mathbf{Y}_i$ is the corresponding clinical report. An ECG encoder $g_\theta: \mathbb{R}^{C \times T} \to \mathbb{R}^d$ and a text encoder $h_\phi: \mathcal{Y} \to \mathbb{R}^d$ produce representations $\mathbf{Z} = g_\theta(\mathbf{X})$ and $\mathbf{R} = h_\phi(\mathbf{Y})$ in a shared latent space of dimension $d$. We collect all trainable parameters as $\Theta = \{\theta, \phi, \psi\}$, where $\psi$ denotes the parameters of an auxiliary decoder introduced below.

\paragraph{Learning Goals.}
A clinically useful ECG representation must satisfy two properties that pull against each other. The first is semantic: $\mathbf{Z}$ should encode the diagnostic concepts that the paired report describes, so that downstream classifiers and retrieval systems can recover clinical labels from the embedding alone. The report itself is a natural target for this requirement. It is written by a clinician at the time of recording, organizes the signal around diagnostically relevant categories (rhythm, axis, conduction abnormalities, ischemic changes), and uses vocabulary that downstream tasks ultimately predict against. Aligning $\mathbf{Z}$ with the text embedding $\mathbf{R}$ therefore pulls the representation toward axes a clinician would consider meaningful, without requiring explicit label supervision. We formalize this as maximizing the mutual information between the two embeddings,
\begin{equation}
\mathrm{I}(\mathbf{Z}; \mathbf{R}) = \int p(\mathbf{z}, \mathbf{r}) \log \frac{p(\mathbf{z}, \mathbf{r})}{p(\mathbf{z})\, p(\mathbf{r})} \, d\mathbf{z} \, d\mathbf{r} \;=\; \mathrm{H}(\mathbf{R}) - \mathrm{H}(\mathbf{R} \mid \mathbf{Z}),
\label{eq:mi_zr}
\end{equation}
which measures how much of the report's content is recoverable from $\mathbf{Z}$: zero when the two embeddings are independent, and equal to $\mathrm{H}(\mathbf{R})$ when $\mathbf{R}$ is a deterministic function of $\mathbf{Z}$.

\begin{figure*}[t]
\centering
\includegraphics[width=0.85\linewidth]{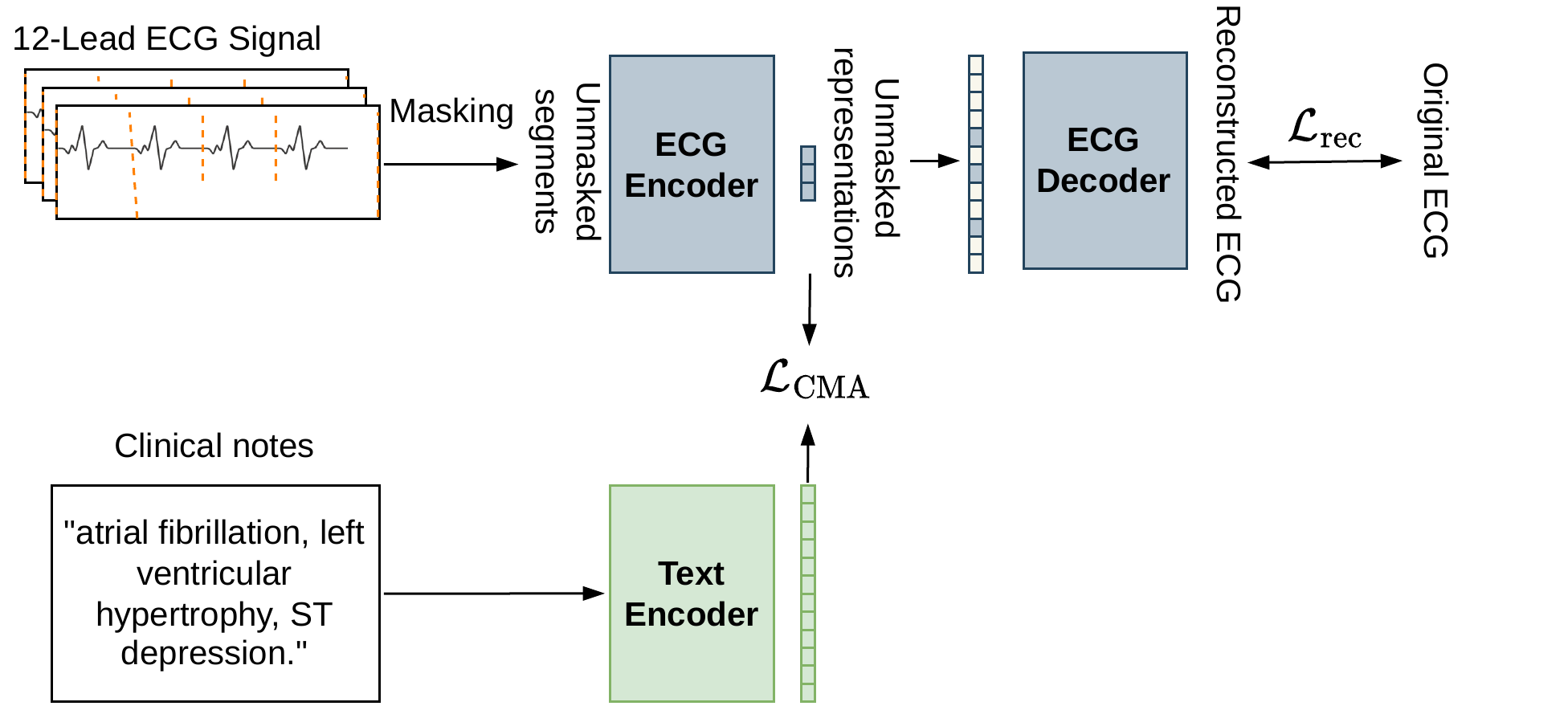}
\label{fig_Infomax}
\caption{
\textbf{Overview of the proposed ECG-text multimodal representation learning framework.} Given a 12-lead ECG signal and corresponding clinical notes, the model learns a shared representation via two complementary objectives. An information maximization (InfoMax) formulation that reconstructs the masked ECG signal using reconstruction loss and aligns ECG and text representations via a cross-modal alignment (CMA) loss.
Together, these objectives promote robust and clinically meaningful multimodal representations.
}
\label{fig:method}
\end{figure*}

The second requirement is structural: $\mathbf{Z}$ should retain the physiological information carried by the raw waveform, including inter-lead correlations, P-QRS-T morphology, beat-to-beat variability, and ST-segment dynamics that are diagnostically meaningful but rarely transcribed in the report \cite{ecg-soup, STMEM2024}. We formalize this analogously as maximizing $\mathrm{I}(\mathbf{Z}; \mathbf{X}) = \mathrm{H}(\mathbf{X}) - \mathrm{H}(\mathbf{X} \mid \mathbf{Z})$, which lower-bounds how faithfully $\mathbf{X}$ can be reconstructed from $\mathbf{Z}$ \cite{hjelm2019learningdeeprepresentationsmutual, lele2025}.

The two objectives are in tension because the report is a heavily compressed summary of the signal: $\mathrm{H}(\mathbf{R}) \ll \mathrm{H}(\mathbf{X})$, and large parts of the signal's entropy correspond to physiological detail the report never mentions. A representation that maximizes $\mathrm{I}(\mathbf{Z}; \mathbf{R})$ alone is free to discard everything in $\mathbf{X}$ that does not appear in $\mathbf{R}$, collapsing $\mathbf{Z}$ onto the report's vocabulary; a representation that maximizes $\mathrm{I}(\mathbf{Z}; \mathbf{X})$ alone preserves signal structure but has no reason to organize it around clinically meaningful axes. Our objective targets both jointly. The remainder of this section develops it: Sec.~\ref{sec:mmim} states the joint objective and derives a tractable lower bound, Sec.~\ref{sec:IB} examines the information-bottleneck alternative and isolates the assumption it makes about the ECG--text setting, and Sec.~\ref{sec:arch} translates the objective into a dual-branch architecture.

\subsection{Multimodal Information Maximization}
\label{sec:mmim}

Following the InfoMax~\cite{lele2025} principle, we jointly maximize the two mutual information identified above:
\begin{equation}
\max_{\theta, \phi} \; \mathrm{I}(\mathbf{Z}; \mathbf{R}) + \mathrm{I}(\mathbf{Z}; \mathbf{X}).
\label{eq:mmim}
\end{equation}
Neither term is directly optimizable: the integrals in Eq.~\eqref{eq:mi_zr} require access to the joint densities $p(\mathbf{z}, \mathbf{r})$ and $p(\mathbf{z}, \mathbf{x})$, which are unavailable in high dimensions. We obtain a tractable objective by replacing each with a standard variational lower bound: a contrastive bound for the cross-modal term and a reconstruction bound for the signal term.

\paragraph{Cross-modal alignment.}
For $\mathrm{I}(\mathbf{Z}; \mathbf{R})$ we use the symmetric InfoNCE bound \cite{oord2019representationlearningcontrastivepredictive}, which lower-bounds mutual information through the ratio between paired and unpaired similarities over a minibatch of size $B$:
\begin{equation}
\mathcal{L}_{\text{CMA}} \;=\; \underbrace{-\frac{1}{B} \sum_{i=1}^{B} \log \frac{\exp(s(\mathbf{Z}_i, \mathbf{R}_i)/\tau)}{\sum_{j=1}^{B} \exp(s(\mathbf{Z}_i, \mathbf{R}_j)/\tau)}}_{\mathcal{L}_{\mathbf{Z} \to \mathbf{R}}} \;+\; \underbrace{-\frac{1}{B} \sum_{i=1}^{B} \log \frac{\exp(s(\mathbf{Z}_i, \mathbf{R}_i)/\tau)}{\sum_{j=1}^{B} \exp(s(\mathbf{Z}_j, \mathbf{R}_i)/\tau)}}_{\mathcal{L}_{\mathbf{R} \to \mathbf{Z}}},
\label{eq:cma}
\end{equation}
where $s(\cdot, \cdot)$ is a similarity function and $\tau$ a temperature. The two terms differ only in which modality indexes the negatives in the denominator: $\mathcal{L}_{\mathbf{Z} \to \mathbf{R}}$ contrasts each ECG against all reports in the batch, $\mathcal{L}_{\mathbf{R} \to \mathbf{Z}}$ contrasts each report against all ECGs. Minimizing $\mathcal{L}_{\text{CMA}}$ tightens a lower bound on $\mathrm{I}(\mathbf{Z}; \mathbf{R})$ that grows with $B$.

\paragraph{Signal preservation.}
For $\mathrm{I}(\mathbf{Z}; \mathbf{X})$ we want a bound that ties preservation of the signal to a quantity the encoder can actually optimize. The natural route is reconstruction: $\mathrm{I}(\mathbf{Z}; \mathbf{X}) = \mathrm{H}(\mathbf{X}) - \mathrm{H}(\mathbf{X} \mid \mathbf{Z})$ is large exactly when $\mathbf{X}$ is nearly determined by $\mathbf{Z}$, so any decoder that can recover $\mathbf{X}$ from $\mathbf{Z}$ certifies that the embedding has retained the signal's content. Following Barber and Agakov \cite{10.5555/2981345.2981371}, we replace the intractable conditional $p(\mathbf{X} \mid \mathbf{Z})$ with a learned decoder $q_\psi(\mathbf{X} \mid \mathbf{Z})$ and use the non-negativity of KL to obtain
\begin{equation}
\mathrm{I}(\mathbf{Z}; \mathbf{X}) \;\geq\; \mathbb{E}_{p(\mathbf{X}, \mathbf{Z})}\!\left[\log q_\psi(\mathbf{X} \mid \mathbf{Z})\right] + \mathrm{H}(\mathbf{X}),
\label{eq:ba_bound}
\end{equation}
with equality when $q_\psi(\mathbf{X} \mid \mathbf{Z}) = p(\mathbf{X} \mid \mathbf{Z})$. The entropy $\mathrm{H}(\mathbf{X})$ is constant in $\Theta$, so maximizing the bound reduces to maximizing the expected log-likelihood under $q_\psi$, and the bound's tightness is governed by how well $q_\psi$ approximates the true posterior.

We instantiate $q_\psi$ as an isotropic Gaussian, $q_\psi(\mathbf{X} \mid \mathbf{Z}) = \mathcal{N}(\mathbf{X}; \mu_\psi(\mathbf{Z}), \sigma^2 \mathbf{I})$ with decoder mean $\mu_\psi(\mathbf{Z})$ and fixed variance $\sigma^2$. This choice is well matched to ECG: the signal is continuous-valued and locally smooth, and after standard preprocessing the residual sensor noise is approximately additive Gaussian, so a Gaussian decoder family contains plausible candidates for $p(\mathbf{X} \mid \mathbf{Z})$ rather than imposing a structurally wrong likelihood. Under this choice, $\log q_\psi$ collapses up to constants to a squared error, giving the reconstruction loss
\begin{equation}
\mathcal{L}_{\text{rec}} = \left\| \mathbf{X} - \mu_\psi(\mathbf{Z}) \right\|_2^2.
\label{eq:rec}
\end{equation}
Minimizing $\mathcal{L}_{\text{rec}}$ tightens the lower bound in Eq.~\eqref{eq:ba_bound} on $\mathrm{I}(\mathbf{Z}; \mathbf{X})$ and pushes $\mathbf{Z}$ to retain the signal content needed to reconstruct $\mathbf{X}$ under the Gaussian decoder family.

\paragraph{Masked reconstruction interpretation.}
In practice, the reconstruction branch operates under masked ECG modeling. Let $\mathbf{X}_{vis}$ denote the visible ECG segments and $\mathbf{X}^{\text{mask}}$ the masked segments, with latent representation $\mathbf{Z} = g_\theta(\mathbf{X}_{vis})$. The decoder predicts the masked regions through a variational distribution $q_\psi(\mathbf{X}^{\text{mask}} \mid \mathbf{Z})$. Under this setting, the reconstruction objective can be interpreted as maximizing a lower bound on the mutual information between the latent representation and the masked signal content:
\begin{equation}
\mathrm{I}(\mathbf{Z}; \mathbf{X}^\text{mask})
\ge
\mathbb{E}_{p(\mathbf{X}^\text{mask}, \mathbf{Z})}
\left[
\log q_\psi(\mathbf{X}^\text{mask} \mid \mathbf{Z})
\right]
+ \mathrm{H}(\mathbf{X}^\text{mask}).
\end{equation}
Thus, minimizing the masked reconstruction loss encourages the representation to preserve predictive physiological structure shared across ECG leads and temporal regions, rather than merely memorizing the observed waveform. 

\paragraph{Final objective.}
Combining the two surrogates yields the training loss
\begin{equation}
\resizebox{\linewidth}{!}{
\colorbox{boxgray}{$\displaystyle
\mathcal{L} \;=\; \underbrace{-\frac{1}{B}\sum_{i=1}^{B} \log \frac{\exp(s(\mathbf{Z}_i, \mathbf{R}_i)/\tau)}{\sum_{j=1}^{B}\exp(s(\mathbf{Z}_i, \mathbf{R}_j)/\tau)} \;-\; \frac{1}{B}\sum_{i=1}^{B} \log \frac{\exp(s(\mathbf{Z}_i, \mathbf{R}_i)/\tau)}{\sum_{j=1}^{B}\exp(s(\mathbf{Z}_j, \mathbf{R}_i)/\tau)}}_{\mathcal{L}_{\text{CMA}}: \text{ semantic alignment with the report}} \;+\; \underbrace{\left\| \mathbf{X} - \mu_\psi(\mathbf{Z}) \right\|_2^2}_{\mathcal{L}_{\text{rec}}: \text{ structural preservation of the signal}}
$}}
\label{eq:final_loss}
\end{equation}
where the cross-modal term is applied to the global pooled representations and the reconstruction term to the masked-modeling branch (Sec.~\ref{sec:arch}). The two terms address the two requirements separately and are optimized jointly without an explicit trade-off coefficient. Thus, minimizing the masked reconstruction loss encourages the latent representation to preserve predictive physiological structure shared across visible and masked ECG regions, rather than merely memorizing the observed waveform. In particular, the encoder is encouraged to capture cross-lead correlations, waveform morphology, and temporal dynamics necessary for predicting missing ECG structure from a partially observed ECG context.

\subsection{An Alternative: Multimodal Information Bottleneck}
\label{sec:IB}

An alternative formulation comes from the Information Bottleneck principle \cite{tishby2000informationbottleneckmethod, tishby2015deeplearninginformationbottleneck}, which casts representation learning as a constrained compression problem: find $\mathbf{Z}$ that retains as much information as possible about a target while discarding as much as possible about the input. Applied to the ECG--text setting with the report as target, this becomes
\begin{equation}
\max_{\theta, \phi} \; \mathrm{I}(\mathbf{Z}; \mathbf{R}) \quad \text{s.t.} \quad \min \, \mathrm{I}(\mathbf{Z}; \mathbf{X}),
\label{eq:ib}
\end{equation}
keeping the cross-modal content the report names and compressing whatever signal structure is not needed to predict it. Recent work applies this principle to image--text alignment \cite{almudévar2025aligningmultimodalrepresentationsinformation}, treating the bottleneck as a way to remove modality-specific noise.

The principle carries an implicit assumption: that the input information being compressed away is, on average, irrelevant to downstream tasks. The entropy comparison from Sec.~\ref{sec:mmim} makes the stakes concrete. Because $\mathrm{H}(\mathbf{R}) \ll \mathrm{H}(\mathbf{X})$, the gap $\mathrm{I}(\mathbf{Z}; \mathbf{X}) - \mathrm{I}(\mathbf{Z}; \mathbf{R})$ that IB compresses away is large in absolute terms, and IB only pays off if that gap is mostly task-irrelevant noise. This holds for image--text settings where captions describe most of what the image depicts, but it is not obvious for ECG--text. Clinical reports summarize a recording at the diagnostic level and routinely omit inter-lead correlations, waveform morphology, and temporal dynamics, features that downstream tasks can depend on \cite{ecg-soup, STMEM2024}. Compressing them away to tighten alignment is therefore a substantive bet about the ECG--text setting, not a neutral regularization choice.

To test this bet directly, we adapt the IB formulation to ECG--text by adding a bottleneck term to the cross-modal objective. Following \cite{almudévar2025aligningmultimodalrepresentationsinformation}, we instantiate the bottleneck as a cosine-alignment penalty between paired ECG and text embeddings:
\begin{equation}
\mathcal{L}_{\text{MIB}} \;=\; \mathcal{L}_{\text{CMA}} \;+\; \lambda_{\text{IB}} \underbrace{\frac{1}{B} \sum_{i=1}^{B} \left(1 - \hat{\mathbf{Z}}_i^{\top} \hat{\mathbf{R}}_i\right)}_{\mathcal{L}_{\text{IB}}},
\label{eq:mib}
\end{equation}
with $\hat{\mathbf{Z}}_i = \mathbf{Z}_i / \|\mathbf{Z}_i\|_2$, $\hat{\mathbf{R}}_i = \mathbf{R}_i / \|\mathbf{R}_i\|_2$, and $\lambda_{\text{IB}} \in \mathbb{R}^{+}$ controlling the bottleneck strength. The choice acts as a practical IB proxy: driving $\hat{\mathbf{Z}}_i$ toward $\hat{\mathbf{R}}_i$ collapses the components of $\mathbf{Z}$ that do not align with $\mathbf{R}$, suppressing modality-specific content of $\mathbf{X}$ in the embedding without explicitly estimating $\mathrm{I}(\mathbf{Z}; \mathbf{X})$.

\subsection{Model Architecture}
\label{sec:arch}

The objective in Eq.~\eqref{eq:final_loss} prescribes the architecture directly: $\mathcal{L}_{\text{rec}}$ requires a path that reconstructs the signal from a latent representation, and $\mathcal{L}_{\text{CMA}}$ requires a path that aligns a global ECG embedding with a text embedding. We instantiate these as two branches sharing the ECG encoder $g_\theta$, illustrated in Fig.~\ref{fig:method}, so that backpropagation forces a single representation to satisfy both losses simultaneously.

\paragraph{Masked ECG modeling branch.}
A stochastic masking operator $M(\cdot)$ produces a partially observed signal $\mathbf{X}^{\text{mask}} = M(\mathbf{X})$, which the encoder maps to visible-token representations $\mathbf{Z}^{\text{vis}} = g_\theta(\mathbf{X}^{\text{mask}})$. The full latent sequence $\mathbf{Z}^{\text{full}}$ is reconstructed by inserting learned mask tokens at the masked positions, and the decoder produces $\mu_\psi(\mathbf{Z}^{\text{full}})$, against which $\mathcal{L}_{\text{rec}}$ is computed. We adopt the STMEM design \cite{STMEM2024} with a lightweight TinyViT backbone, which captures local waveform structure and long-range temporal dependencies at a parameter budget compatible with joint training.

\paragraph{Alignment branch.}
The visible representations $\mathbf{Z}^{\text{vis}}$ are aggregated by average pooling into a single ECG embedding $\mathbf{Z}$. The clinical report $\mathbf{Y}$ is encoded by the pretrained MedCPT model $h_\phi(\cdot)$ \cite{Jin_2023} and similarly pooled into a text embedding $\mathbf{R}$. Both are passed through projection heads into a shared space, yielding $\tilde{\mathbf{Z}}$ and $\tilde{\mathbf{R}}$, on which $\mathcal{L}_{\text{CMA}}$ is computed. Sharing $g_\theta$ across branches couples the objectives: $\mathcal{L}_{\text{rec}}$ shapes the encoder to preserve signal content while $\mathcal{L}_{\text{CMA}}$ shapes it to surface clinical semantics, both acting on the same underlying parameters.

\section{Experiments}

\subsection{Implementation Details}
\label{sec:implementation}

\paragraph{Pre-training.}
We pre-train on MIMIC-ECG \cite{PhysioNet-mimic-iv-ecg-1.0}, which contains 800,035 ECG--report pairs from 161,352 subjects, each consisting of a 12-lead ECG signal and a corresponding clinical report. We construct the text input by concatenating the diagnostic notes of each report into a single textual description, providing a consistent target for multimodal alignment. The ECG encoder $g_\theta$ is a one-dimensional TinyViT, and the text encoder $h_\phi$ is the pretrained MedCPT model \cite{Jin_2023}.

\paragraph{Downstream evaluation.}
We evaluate the pre-trained model on three public ECG benchmarks covering diverse multi-label classification tasks: PTB-XL \cite{PhysioNet-ptb-xl-1.0.3, patrick2020}, CPSC \cite{Ng2018AnOA}, and Chapman--Shaoxing--Ningbo (CSN) \cite{zheng2020, PhysioNet-ecg-arrhythmia-1.0.0}. We compare against both unimodal SSL and multimodal representation learning methods across four complementary settings: linear probing, which assesses representation quality under a frozen encoder; zero-shot classification, which assesses cross-modal generalization without task-specific fine-tuning; domain-shift evaluation, which assesses robustness across datasets with aligned label spaces; and text generation, which assesses whether the learned representations support language-conditioned downstream tasks. For ECG-conditioned text generation, we also use the ECG Question-Answer (ECG-QA) dataset \cite{lai2025medr1}, which contains ECG waveforms paired with clinically grounded question-and-answer-style textual interpretations. Preprocessing, hyperparameters, and dataset details are provided in App.~\ref{implementation_details}.

\subsection{Results \& Discussion}
\subsubsection{ECG Linear Probing}
\begin{table*}[t]
\centering
\scriptsize
\setlength{\tabcolsep}{0.98pt}
\caption{\textbf{Performance comparison} across multiple label taxonomies (PTB-XL SuperClass, SubClass, Form, Rhythm, and All conditions, CSN, and CPSC). Best results are in bold and second-best shaded in gray. Detailed results, including standard deviations, are reported in Table \ref{tab:ptbxl_results_detail} in Appendix \ref{extended_eval}.}
\label{tab:ptbxl_results}
\renewcommand{\arraystretch}{0.76}
\begin{tabular}{lccccccccccccccc}
\toprule
& \multicolumn{3}{c}{PTB-XL SuperClass} 
& \multicolumn{3}{c}{PTB-XL SubClass} 
& \multicolumn{3}{c}{PTB-XL Form}
& \multicolumn{3}{c}{PTB-XL Rhythm}
& \multicolumn{3}{c}{PTB-XL All} \\
\cmidrule(lr){2-4} \cmidrule(lr){5-7} \cmidrule(lr){8-10} \cmidrule(lr){11-13} \cmidrule(lr){14-16}
Methods 
& AUC & F1 & Accuracy
& AUC & F1 & Accuracy
& AUC & F1 & Accuracy
& AUC & F1 & Accuracy
& AUC & F1 & Accuracy \\
\midrule
ECG-FM \cite{ecg-fm2025}
& $87.58$ & $67.21$ & $84.10$
& $87.23$ & $41.68$ & $92.41$
& $78.19$ & $31.06$ & $85.26$
& $88.50$ & $54.41$ & $95.76$
& $83.87$ & $31.31$ & $92.26$ \\

STMEM \cite{STMEM2024} 
& $92.03$ & $73.52$ & $87.52$
& $91.74$ & $\cellcolor{gray!30} 49.82$ & $94.32$
& $\textbf{89.15}$ & $\textbf{45.68}$ & $\textbf{92.24}$
& $\cellcolor{gray!30} 98.02$ & $\textbf{68.55}$ & $\cellcolor{gray!30} 98.61$
& $91.02$ & $39.30$ & $95.22$ \\

MERL \cite{merl2024} 
& $91.20$ & $72.69$ & $87.19$
& $92.70$ & $49.54$ & $\cellcolor{gray!30} 94.95$
& $83.72$ & $36.47$ & $88.99$
& $91.03$ & $53.27$ & $95.55$
& $88.97$ & $35.80$ & $93.88$ \\

ESI \cite{esi2024}
& $85.98$ & $65.48$ & $82.98$
& $83.34$ & $33.68$ & $89.41$
& $77.34$ & $30.81$ & $84.19$
& $90.37$ & $39.69$ & $95.12$
& $74.81$ & $18.22$ & $84.19$ \\

D-BETA \cite{dbeta2025}
& $91.21$ & $72.56$ & $87.26$
& $91.66$ & $48.93$ & $94.94$
& $85.23$ & $39.16$ & $90.68$
& $97.55$ & $65.77$ & $98.37$
& $\cellcolor{gray!30} 91.52$ & $\cellcolor{gray!30} 39.34$ & $\cellcolor{gray!30} 95.53$ \\

QoQ \cite{qoq2025}  
& $88.54$ & $68.80$ & $85.28$
& $87.96$ & $41.50$ & $93.42$
& $81.01$ & $31.78$ & $87.29$
& $88.41$ & $46.41$ & $94.95$
& $83.76$ & $28.55$ & $92.40$ \\

ECG-Chat \cite{ecg-chat2025}  
& $\cellcolor{gray!30} 92.25$ & $\cellcolor{gray!30} 74.40$ & $\cellcolor{gray!30} 87.96$
& $\cellcolor{gray!30} 91.74$ & $48.98$ & $94.94$ 
& $84.17$ & $39.58$ & $89.32$
& $94.25$ & $58.76$ & $97.80$
& $88.01$ & $36.40$ & $94.42$ \\
\midrule
Our 
& $\textbf{93.26}$ & $\textbf{75.97}$ & $\textbf{89.03}$
& $\textbf{93.79}$ & $\textbf{54.89}$ & $\textbf{95.68}$
& $\cellcolor{gray!30} 87.76$ & $\cellcolor{gray!30} 44.14$ & $\cellcolor{gray!30} 91.54$
& $\textbf{98.03}$ & $\cellcolor{gray!30} 67.50$ & $\textbf{98.63}$
& $\textbf{92.70}$ & $\textbf{42.63}$ & $\textbf{95.80}$ \\
\bottomrule
\end{tabular}


\begin{tabular}{lcccccc}
\toprule
& \multicolumn{3}{c}{CSN}
& \multicolumn{3}{c}{CPSC} \\
\cmidrule(lr){2-4} \cmidrule(lr){5-7}
Methods
& AUC & F1 & Accuracy
& AUC & F1 & Accuracy \\
\midrule
ECG-FM \cite{ecg-fm2025}
& $88.70$ & $41.33$ & $96.26$
& $90.55$ & $67.18$ & $92.49$  \\

STMEM \cite{STMEM2024}
& $\cellcolor{gray!30} 95.35$ & $\cellcolor{gray!30} 52.83$ & $\cellcolor{gray!30} 97.52$
& $96.36$ & $79.96$ & $95.92$ \\

MERL \cite{merl2024}
& $91.32$ & $43.95$ & $95.79$ 
& $90.65$ & $66.84$ & $91.52$ \\

ESI \cite{esi2024}
& $77.01$ & $29.68$ & $85.28$ 
& $92.84$ & $70.99$ & $93.86$ \\

D-BETA \cite{dbeta2025}
& $95.16$ & $52.13$ & $97.49$ 
& $\textbf{96.51}$ & $\textbf{81.37}$ & $\textbf{96.20}$ \\

QoQ \cite{qoq2025}
& $85.41$ & $35.14$ & $93.04$
& $88.98$ & $63.23$ & $90.41$ \\

ECG-Chat \cite{ecg-chat2025}  
& $93.83$ & $50.42$ & $97.15$ 
& $95.06$ & $76.94$ & $95.28$ \\
\midrule
Our 
& $\textbf{96.46}$ & $\textbf{55.55}$ & $\textbf{97.82}$
& $\cellcolor{gray!30} 96.46$ & $\cellcolor{gray!30} 79.88$ & $\cellcolor{gray!30} 95.93$ \\
\bottomrule
\end{tabular}

\end{table*}

Table \ref{tab:ptbxl_results} presents the linear probing results across multiple label taxonomies on PTB-XL, CSN, and CPSC. We compare against both unimodal ECG SSL methods (ECG-FM and STMEM), multimodal ECG-text alignment frameworks (MERL, ESI, D-BETA), and ECG-text generation-based approaches (QoQ and ECG-Chat). Our method consistently achieves the strongest overall performance, particularly in AUC and Accuracy. In the few cases where it does not rank first, it consistently achieves second-best results, indicating strong and stable performance across diverse settings.

A key observation is the consistent improvement on fine-grained tasks, including PTB-XL Sub-class, PTB-XL All, and CSN, where our method outperforms all baselines by a clear margin. These settings require capturing subtle physiological variations, suggesting that the model better preserves intrinsic signal-specific and clinically relevant characteristics. In contrast, in general-label settings such as PTB-XL Superclass, Rhythm, and CPSC, our method remains highly competitive and always achieves the best or second-best performance. Together, these results indicate that the proposed method is effective across multiple levels of ECG abstraction, from coarse diagnostic categories (Superclass, Rhythm) to fine-grained morphological and hierarchical labels (Subclass, Form, and All), without introducing the trade-off typically observed in alignment-dominant methods.  From an information-theoretic perspective, these results highlight that jointly maximizing $\mathrm{I}(\mathbf{Z}; \mathbf{X})$ and $\mathrm{I}(\mathbf{Z}; \mathbf{R})$ enables the model to preserve intrinsic physiological structure while effectively integrating clinical knowledge into the ECG representation. As a result, even when using only the ECG encoder, the learned representations outperform both unimodal and prior multimodal approaches, demonstrating superior expressiveness and generalization.


\subsubsection{Zero-shot Evaluation}
Table \ref{tab:zeroshot_results} reports the zero-shot evaluation effects on PTB-XL under multiple label taxonomies. Unlike linear probing, zero-shot evaluation requires both ECG and text representations. In this context, unimodal ECG foundation models such as STMEM and ECG-FM are not applicable, as they do not incorporate textual representations. Similarly, generative approaches such as QoQ, which rely on LLM-based decoding rather than shared embedding alignment, are not directly comparable in zero-shot classification. Also in this setting, our method consistently performs strongly across most label groups. It frequently ranks among the top-performing methods, with multiple best- and second-best results across different taxonomies, indicating stable, well-balanced performance. Notably, the model achieves the best performance on SuperClass AUC ($76.59\%$) and SubClass metrics (AUC: $77.58\%$), highlighting its ability to capture both global diagnostic information and fine-grained physiological patterns. Strong performance is also observed in the Form setting, suggesting enhanced sensitivity to morphological characteristics and representation robustness across different levels of abstraction. Additional evaluations on external datasets (CPSC and CSN) confirm consistent validity, with competitive or superior performance as compared to existing baselines. These results highlight the robustness and transferability of the pre-trained representations for zero-shot ECG classification.

\begin{table*}[t]
\centering
\scriptsize
\setlength{\tabcolsep}{0.98pt}
\renewcommand{\arraystretch}{0.78}
\caption{\textbf{Zero-shot evaluation results across datasets.} Results are reported as mean percentages, with the best results highlighted in bold and the second-best shaded in gray.}
\label{tab:zeroshot_results}
\begin{tabular}{lccccccccccccccc}
\toprule
& \multicolumn{3}{c}{PTB-XL SuperClass} 
& \multicolumn{3}{c}{PTB-XL SubClass} 
& \multicolumn{3}{c}{PTB-XL Form}
& \multicolumn{3}{c}{PTB-XL Rhythm}
& \multicolumn{3}{c}{PTB-XL All} \\
\cmidrule(lr){2-4} \cmidrule(lr){5-7} \cmidrule(lr){8-10} \cmidrule(lr){11-13} \cmidrule(lr){14-16}
Methods 
& AUC & F1 & Accuracy 
& AUC & F1 & Accuracy 
& AUC & F1 & Accuracy
& AUC & F1 & Accuracy
& AUC & F1 & Accuracy \\
\midrule

MERL \cite{merl2024} 
& \cellcolor{gray!30} 74.29 & \cellcolor{gray!30} 52.92 & \cellcolor{gray!30} 69.15
& 74.15 & 25.43 & 84.86
& 64.15 & 20.17 & \textbf{71.98} 
& 78.61 & 26.70 & 84.84
& 70.92 & 16.50 & 83.45 \\

ESI \cite{esi2024} 
& 67.00 & 49.17 & 57.25
& \cellcolor{gray!30} 74.92 & 24.34 & \cellcolor{gray!30} 86.42
& \cellcolor{gray!30} 65.07 & 20.48 & \cellcolor{gray!30} 67.34
& \cellcolor{gray!30} 82.96 & 28.05 & \cellcolor{gray!30} 93.65
& \textbf{76.05} & 17.72 & \cellcolor{gray!30} 87.63 \\

D-BETA \cite{dbeta2025}
& 71.70 & 50.69 & 68.15
& 68.75 & 18.16 & 74.84
& 60.52 & 17.99 & 58.10
& \textbf{91.01} & \textbf{37.03} & \textbf{93.69}
& 72.73 & 15.00 & 79.97 \\

ECG-Chat \cite{ecg-chat2025}
& 64.20 & 51.81 & 67.21
& 73.28 & \cellcolor{gray!30} 26.84 & 83.48
& 65.02 & \cellcolor{gray!30} 21.08 & 66.34
& 81.69 & 29.88 & 93.39
& 73.40 & \cellcolor{gray!30} 18.41 & \textbf{87.91} \\

\textbf{Our} 
& \textbf{76.59} & \textbf{55.07} & \textbf{73.80}
& \textbf{77.58} & \textbf{28.95} & \textbf{89.96}
& \textbf{66.89} & \textbf{24.17} & 64.68
& 82.76 & \cellcolor{gray!30} 32.96 & 90.48
& \cellcolor{gray!30} 74.73 & \textbf{20.37} & 87.07 \\

\bottomrule
\end{tabular}


\begin{tabular}{lcccccc}
\toprule
& \multicolumn{3}{c}{CSN} 
& \multicolumn{3}{c}{CPSC} \\
\cmidrule(lr){2-4} \cmidrule(lr){5-7}
Methods 
& AUC & F1 & Accuracy 
& AUC & F1 & Accuracy \\
\midrule

MERL \cite{merl2024} 
& 74.17 & 21.92 & 83.79
& 76.40 & 39.62 & 77.87 \\

ESI \cite{esi2024}
& 78.30 & 22.68 & \cellcolor{gray!30} 89.84
& 77.61 & 44.75 & 76.40 \\

D-BETA \cite{dbeta2025}
& 73.65 & 17.34 & 80.63
& 77.73 & 37.68 & 75.39 \\

ECG-Chat \cite{ecg-chat2025}
& \textbf{83.05} & \textbf{32.31} & \textbf{91.23}
& \cellcolor{gray!30} 79.42 & \cellcolor{gray!30} 50.51 & \cellcolor{gray!30} 78.50 \\

\textbf{Our} 
& \cellcolor{gray!30} 78.49 & \cellcolor{gray!30} 24.86 & 89.18
& \textbf{83.20} & \textbf{52.18} & \textbf{86.45} \\
\bottomrule
\end{tabular}
\end{table*}

\subsubsection{Distribution-shift Evaluation}
We further evaluate cross-domain robustness under distribution shift, with detailed settings provided in the Appendix \ref{extended_eval}. As shown in Table \ref{tab:cross_domain}, our method achieves strong performance across transfer settings, indicating more transferable ECG representations than unimodal or alignment-only baselines. Our method outperforms in two transfer directions, from CPSC to CSN and from CSN to PTB-XL Super, and achieves second-best performance in several other settings, including transfers from PTB-XL Super to CSN and from CSN to CPSC. This suggests that the learned representations capture ECG patterns that generalize across cohorts, acquisition protocols, and label spaces, rather than being optimized for a single setting.












\subsubsection{Text Generation Evaluation}

\begin{figure}[H]
    \centering
    \includegraphics[width=\linewidth]{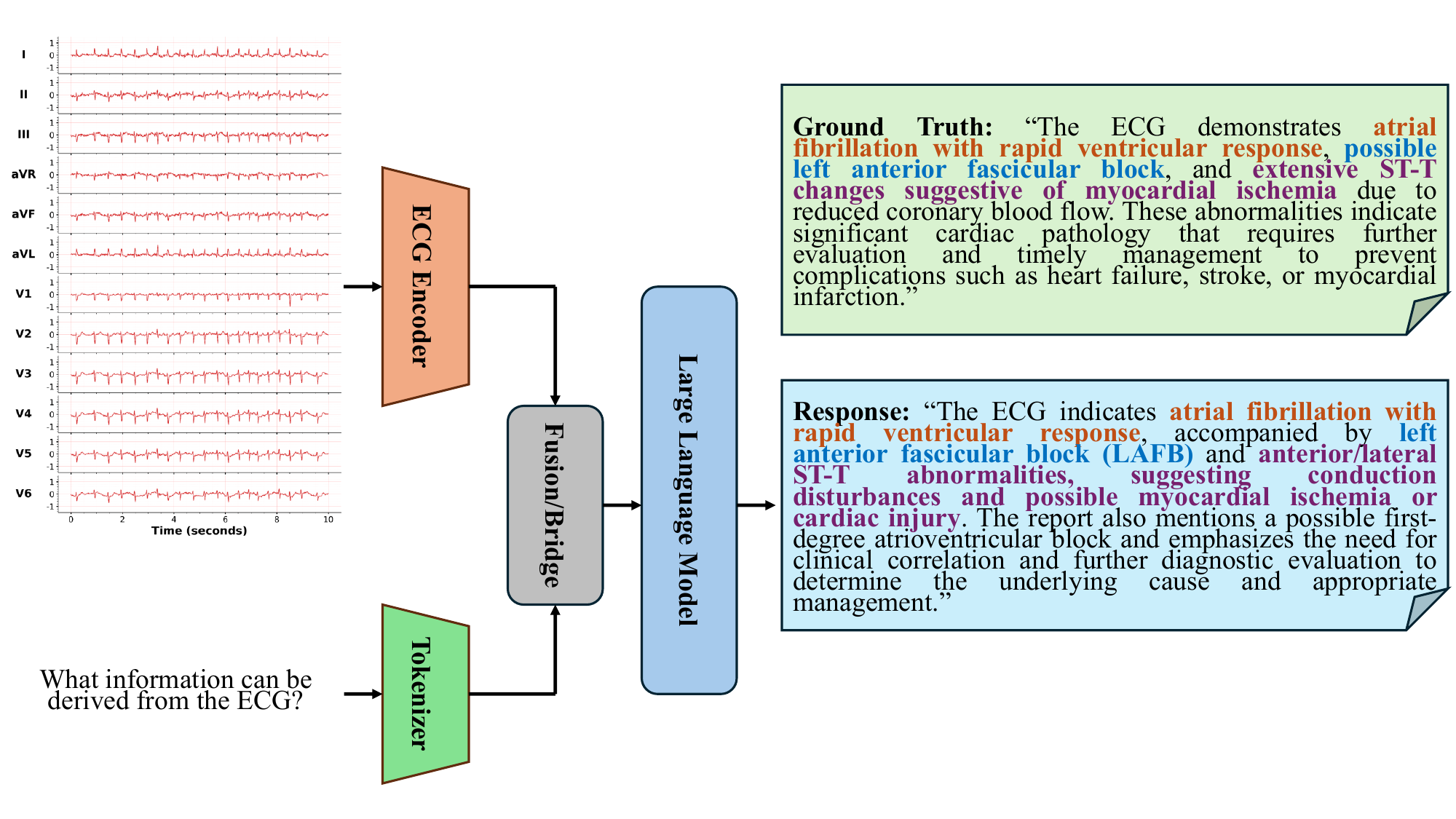}
    \caption{
    \textbf{Illustration of ECG-conditioned clinical text generation.}
    ECG representations extracted from the pretrained ECG encoder are projected into the LLM via a fusion/bridge module following MedTVT-R1~\cite{lai2025medr1} and used as conditioning signals for clinically grounded text generation. Matching colors indicate semantically aligned clinical findings between the generated response and the reference report. Displayed texts are shortened for readability and visualization purposes while preserving their clinical meaning; full reports are provided in the Appendix \ref{downsteam_datasets}. 
    }
    \label{fig:text_generation_example}
\end{figure}

We additionally evaluate the quality of the learned ECG representations in a completely different task: clinical text generation. Inspired by recent multimodal LLM frameworks \cite{lai2025medr1}, we augment the pretrained ECG encoder with an LLM (LLaMA3.2-1B-Instruct) to generate ECG-conditioned text. Figure~\ref{fig:text_generation_example} illustrates the overall ECG-conditioned clinical text generation framework used in our evaluation. During training, the ECG encoder remains frozen, while a lightweight projection adapter maps ECG latent representations into the language model's token embedding space. LoRA modules are further applied to LLaMA3.2-1B-Instruct for parameter-efficient adaptation. Detailed implementation settings are provided in the Appendix \ref{downsteam_datasets}.   After using the pretrained encoders of various models as inputs for subsequent text generation, the quality of the generated text can be evaluated using metrics such as BLEU, METEOR, ROUGE, and BERTScore. The quality metrics are reported in Table \ref{tab:text_generation} for various encoders. Compared with the stage-1 setting of MedTVT-R1, our method improves BLEU from $10.91\%$ to $12.16\%$, METEOR from $32.37\%$ to $33.35\%$, ROUGE from $24.35\%$ to $25.24\%$, and BERTScore from $86.93\%$ to $87.11\%$.

Improved text quality generation based on our ECG representations suggests that the proposed MERIT objective leads to richer, more clinically informative representations. Since all methods share the same LLM architecture and training protocol, the improvements mainly stem from the quality of the learned ECG representation. This observation strongly supports the central hypothesis that joint preservation of signal-specific physiological structure and cross-modal semantic information leads to more transferable and semantically informative ECG representations.

\begin{table}[t]
\centering
\scriptsize
\caption{Text generation evaluation on the MedTVT-R1 QA test set. We adopt the same LLM backbone and stage-1 training pipeline from MedTVT-R1 while replacing only the ECG encoder. Results are reported using BLEU, METEOR, ROUGE, and BERTScore.}
\label{tab:text_generation}
\resizebox{0.60\textwidth}{!}{
\begin{tabular}{lcccc}
\toprule
Method & BLEU & METEOR & ROUGE & BERTScore \\
\midrule
ECG-Chat \cite{ecg-chat2025} & $10.67$ & $32.70$ & $24.37$ & $86.71$ \\
QoQ \cite{qoq2025} & $10.04$ & $31.14$ & $23.29$ & $86.40$ \\
MedTVT-R1 \cite{lai2025medr1} & 10.91 & 32.37 & 24.35 & 86.93 \\
\textbf{Ours} & $\textbf{12.16}$ & $\textbf{33.35}$ & $\textbf{25.24}$ & $\textbf{87.11}$ \\
\bottomrule
\end{tabular}
}
\end{table}

\subsection{Ablation Study}
\begin{figure}[t]
    \centering

    \begin{subfigure}[c]{0.45\linewidth}
        \centering
        \includegraphics[width=\linewidth]{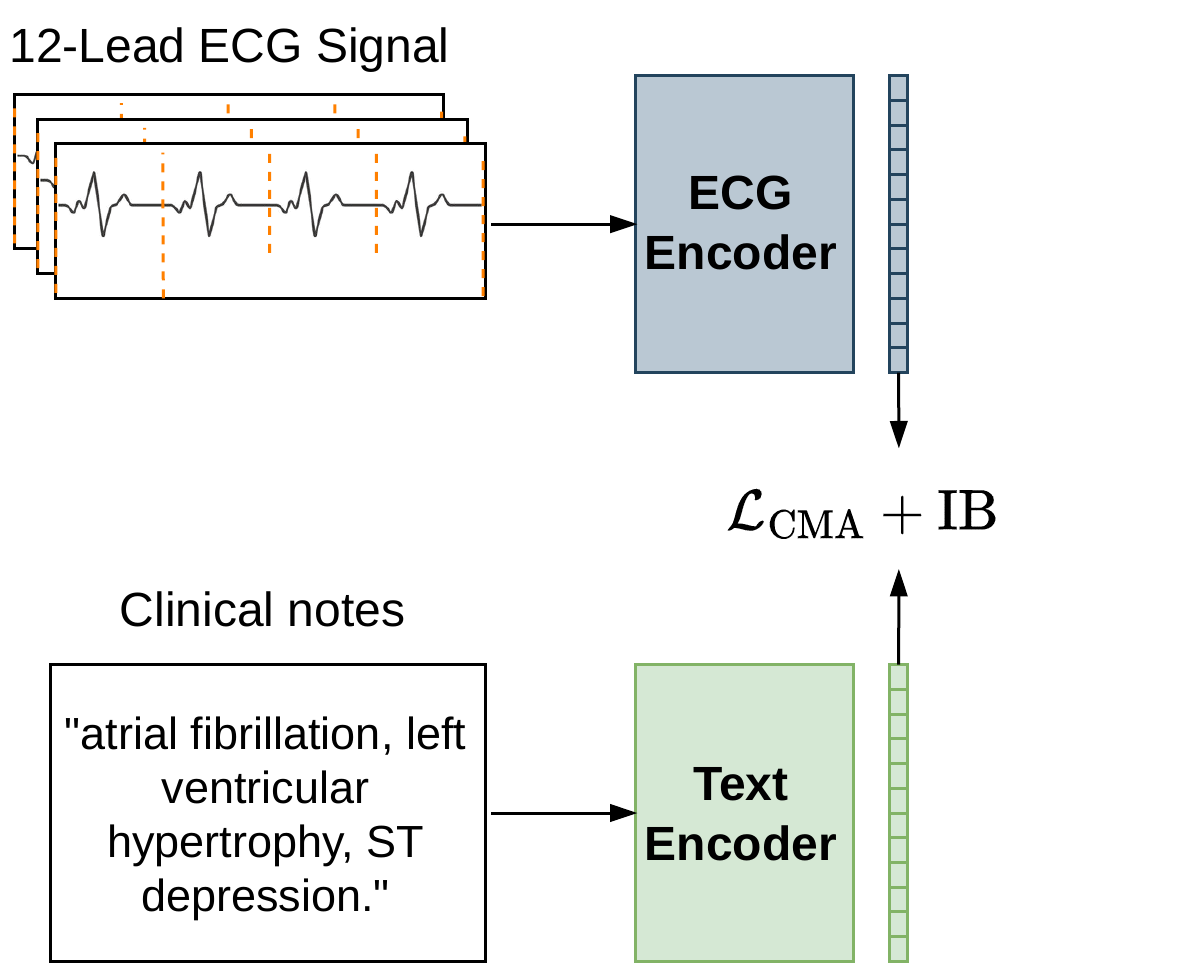}
        \caption{An IB formulation that enforces compact and informative representations while aligning ECG and text embeddings.}
    \label{IB_regularization}
    \end{subfigure}
    \hfill
    \begin{subfigure}[c]{0.54\linewidth}
        \centering

        \begin{subfigure}[t]{0.48\linewidth}
            \centering
            \includegraphics[width=\linewidth]{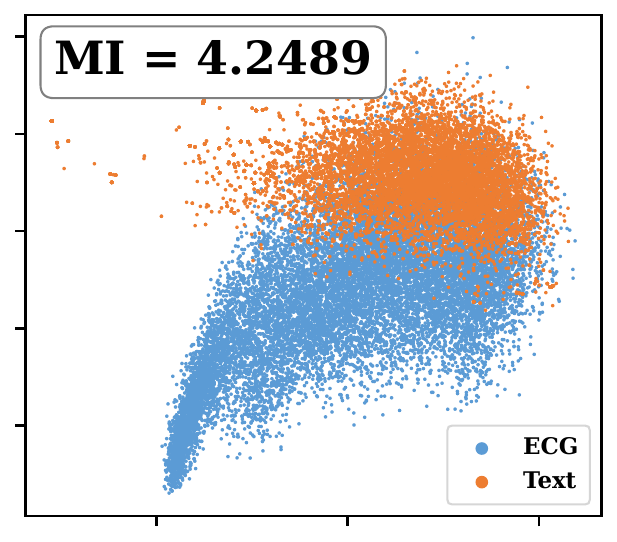}
            \caption{Our method.}
        \end{subfigure}
        \begin{subfigure}[t]{0.48\linewidth}
            \centering
            \includegraphics[width=\linewidth]{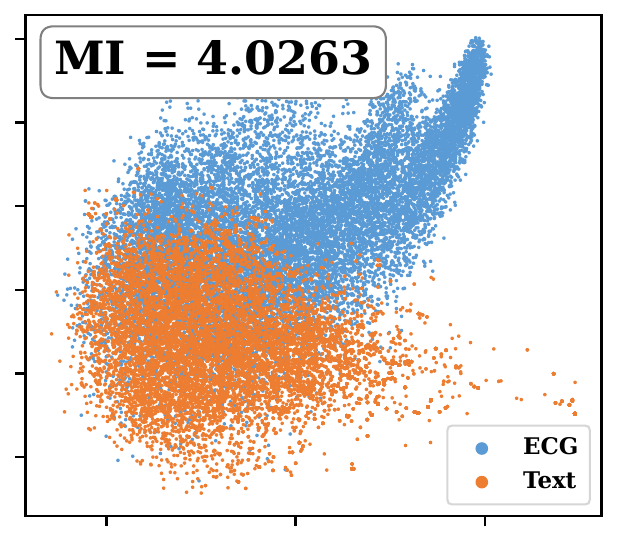}
            \caption{CMA.}
        \end{subfigure}


        \begin{subfigure}[t]{0.48\linewidth}
            \centering
            \includegraphics[width=\linewidth]{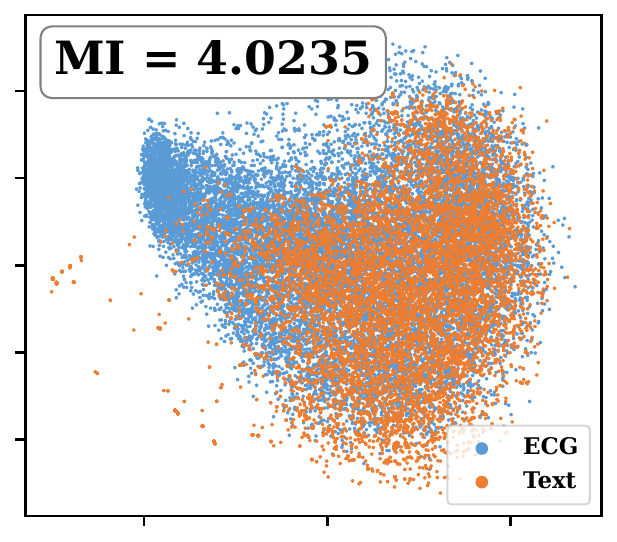}
            \caption{CMA with IB, $\lambda = 0.1$.}
        \end{subfigure}
        \begin{subfigure}[t]{0.48\linewidth}
            \centering
            \includegraphics[width=\linewidth]{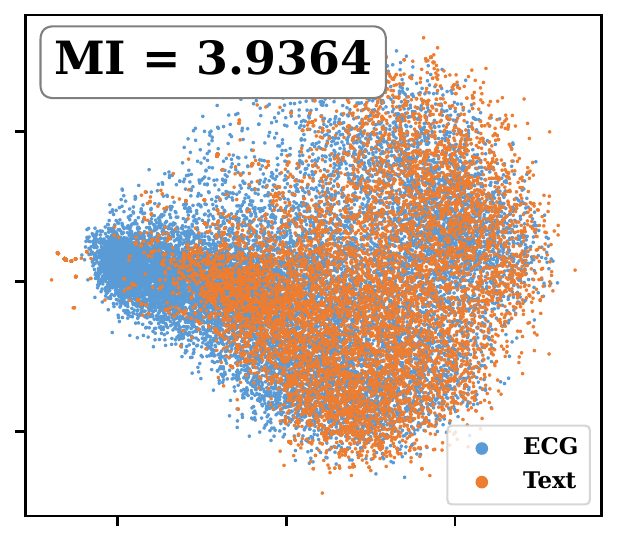}
            \caption{CMA with IB, $\lambda = 1$.}
        \end{subfigure}

    \end{subfigure}

    \caption{\textbf{ECG–Text embedding alignment across methods.} Visualization of the shared representation space on the MIMIC validation set (15{,}223 ECG–Text pairs). Our method achieves strong cross-modal alignment while maintaining ECG representations with richer ECG modality-specific information and higher mutual information (MI) between ECG and text embeddings. CMA also preserves some unique ECG information, but to a lesser extent than our approach. In contrast, IB-based approaches progressively compress unique ECG information as the bottleneck strength increases.}
    
    \label{ecg_text_space}
\end{figure}

The ablation results in Table~\ref{tab:ablation_linear} and \ref{tab:ablation_zeroshot} provide insights into the role of each component. We consider three variants: (i) reconstruction-only training using the mean square error (MSE) objective, (ii) cross-modal alignment-only training using CMA, and (iii) CMA with IB regularization~\cite{almudévar2025aligningmultimodalrepresentationsinformation} as illustrated in Section \ref{sec:IB} and Fig. \ref{IB_regularization}. The proposed MERIT framework jointly optimizes reconstruction and CMA.  Overall, it consistently achieves the best performance across linear probing settings, while maintaining the best or second-best performance in zero-shot evaluation. In linear probing, where performance depends entirely on the quality of the learned ECG representation, the CMA variant generally outperforms IB and often ranks second-best overall. This suggests that the IB objective tends to suppress modality-specific physiological structure by enforcing stronger alignment between ECG and text. Such signal-specific information is particularly important for ECG, where spatio-temporal dependencies, inter-lead correlations, and hierarchical label structures (e.g., superclass, subclass, form, and fine-grained conditions) play a critical role. In contrast, IB achieves stronger performance than CMA in most of zero-shot settings due to its stronger emphasis on cross-modal alignment. In some coarse-grained settings (e.g., PTB-XL SuperClass and CPSC), it even slightly surpasses our method, reflecting its bias toward shared high-level representations while reducing ECG-specific information. Nevertheless, the proposed MERIT framework maintains the strongest overall balance between structural preservation and semantic alignment across both evaluation settings.

To further analyze the learned representations, we visualize the embedding space in Fig.~\ref{ecg_text_space}. The visualization shows that ECG embeddings span a broader region that largely encompasses the text embeddings, reflecting the intrinsic asymmetry between modalities: ECG signals contain substantially richer spatio-temporal and physiological information than clinical text. This observation supports our design choice of integrating clinical semantics into representations that preserve rich ECG structure rather than enforcing strict alignment. Compared with CMA and IB, our method also retains a larger region of ECG-specific information while achieving higher mutual information between ECG and text embeddings. In contrast, IB progressively compresses ECG-specific information as the bottleneck strength increases, consistent with its objective of suppressing modality-specific structure. Figure~\ref{fig:umap} (App. \ref{extended_eval}) provides a complementary view through UMAP visualization on the PTB-XL Rhythm dataset. Compared with CMA and IB variants, our method produces more compact and better-separated clusters across the 12 rhythm classes. In particular, the rare rhythm class PACE forms a more distinguishable cluster under the proposed framework. Together, these results suggest that the proposed MERIT framework not only improves ECG-text semantic alignment while preserving clinically meaningful physiological structure.

\begin{table*}[t]
\centering
\scriptsize
\setlength{\tabcolsep}{0.98pt}
\renewcommand{\arraystretch}{0.76}
\caption{Ablation study under linear probing on PTB-XL, CSN, and CPSC. Detailed results, including standard deviations, are reported in Table \ref{tab:ablation_linear_detail} in Appendix \ref{extended_eval}.}
\label{tab:ablation_linear}

\begin{tabular}{@{}lccc@{\hspace{0.2cm}}ccc@{\hspace{0.2cm}}ccc@{\hspace{0.2cm}}ccc@{\hspace{0.2cm}}ccc@{}}
\toprule
& \multicolumn{3}{c}{PTB-XL SuperClass} & \multicolumn{3}{c}{PTB-XL SubClass} & \multicolumn{3}{c}{PTB-XL Form} & \multicolumn{3}{c}{PTB-XL Rhythm} & \multicolumn{3}{c}{PTB-XL All}\\
\cmidrule(lr){2-4} \cmidrule(lr){5-7} \cmidrule(lr){8-10} \cmidrule(lr){11-13} \cmidrule(lr){14-16}
Methods & AUC & F1 & Accuracy & AUC & F1 & Accuracy & AUC & F1 & Accuracy & AUC & F1 & Accuracy & AUC & F1 & Accuracy \\
\midrule
MSE & $89.52$ & $69.58$ & $85.63$
    & $88.98$ & $44.32$ & $92.89$ 
    & $84.83$ & $37.23$ & $90.41$
    & $95.94$ & $60.93$ & $\cellcolor{gray!30} 98.16$
    & $88.25$ & $34.04$ & $94.11$ \\
IB  & $92.59$ & $74.91$ & $88.56$
    & $92.92$ & $\cellcolor{gray!30} 52.77$ & $\cellcolor{gray!30} 95.20$ 
    & $\cellcolor{gray!30} 86.77$ & $40.66$ & $\cellcolor{gray!30} 90.63$
    & $96.81$ & $\cellcolor{gray!30} 63.74$ & $98.11$
    & $91.67$ & $40.33$ & $95.23$ \\

CMA & $\cellcolor{gray!30} 92.67$ & $\cellcolor{gray!30} 75.17$ & $\cellcolor{gray!30} 88.69$
    & $\cellcolor{gray!30} 93.01$ & $52.46$ & $95.11$ 
    & $86.34$ & $\cellcolor{gray!30} 41.16$ & $89.87$
    & $\cellcolor{gray!30} 96.89$ & $62.92$ & $98.09$
    & $\cellcolor{gray!30} 91.67$ & $\cellcolor{gray!30} 40.35$ & $\cellcolor{gray!30} 95.36$ \\
Our & $\textbf{93.26}$ & $\textbf{75.97}$ & $\textbf{89.03}$
    & $\textbf{93.79}$ & $\textbf{54.89}$ & $\textbf{95.68}$
    & $\textbf{87.76}$ & $\textbf{44.14}$ & $\textbf{91.54}$
    & $\textbf{98.03}$ & $\textbf{67.50}$ & $\textbf{98.63}$
    & $\textbf{92.70}$ & $\textbf{42.63}$ & $\textbf{95.80}$ \\
\bottomrule
\end{tabular}


\begin{tabular}{@{}lccc@{\hspace{0.5cm}}ccc@{}}
\toprule
& \multicolumn{3}{c}{CSN} & \multicolumn{3}{c}{CPSC} \\
\cmidrule(lr){2-4} \cmidrule(lr){5-7}
Methods & AUC & F1 & Accuracy & AUC & F1 & Accuracy \\
\midrule
MSE & $92.68$ & $42.97$ & $96.33$
    & $93.82$ & $73.03$ & $94.15$ \\
IB  & $95.39$ & $\cellcolor{gray!30} 51.44$ & $97.31$
    & $94.86$ & $75.30$ & $94.71$ \\
CMA & $\cellcolor{gray!30} 95.48$ & $51.33$ & $\cellcolor{gray!30} 97.40$
    & $\cellcolor{gray!30} 95.29$ & $\cellcolor{gray!30} 76.46$ & $\cellcolor{gray!30} 95.20$ \\
Our & $\textbf{96.46}$ & $\textbf{55.55}$ & $\textbf{97.82}$
    & $\textbf{96.46}$ & $\textbf{79.88}$ & $\textbf{95.93}$ \\
\bottomrule
\end{tabular}
\end{table*}

\section{Conclusion}
We proposed MERIT, a principled framework for multimodal ECG representation learning that jointly preserves ECG-specific physiological structure and aligns representations with clinical semantics. By formulating multimodal learning from an information-theoretic perspective, our method unifies signal reconstruction and cross-modal alignment within a single objective. Extensive experiments demonstrate consistent improvements across linear probing, zero-shot, cross-domain transfer, and ECG-conditioned clinical text generation settings, particularly for fine-grained clinical tasks. Representation analyses further show that the proposed framework preserves richer ECG-specific information while maintaining effective ECG-text alignment. These findings suggest that balancing structural preservation and semantic integration is critical for learning expressive and generalizable multimodal ECG representations.


\newpage

\bibliographystyle{unsrtnat}
\FloatBarrier
\bibliography{references}

@article{muhammad2023,
title={Global {ECG} Classification by Self-Operational Neural Networks With Feature Injection},
author={Muhammad Uzair Zahid and Serkan Kiranyaz and Moncef Gabbouj},
journal={IEEE Transactions on Biomedical Engineering},
volume={70},
number={1},
pages={205 - 215},
year={2023},
publisher={IEEE}
}

@article{khiem2023,
title={Light{X3ECG}: A Lightweight and eXplainable Deep Learning System for 3-lead Electrocardiogram Classification},
author={Khiem H. Le and others},
journal={Biomedical Signal Processing and Control},
volume={85},
year={2023},
publisher={Elsevier}
}

@article{shengnan2023,
title={G2-Res{N}e{X}t: A Novel Model for {ECG} Signal Classification},
author={Shengnan Hao and others},
journal={IEEE Access},
volume={11},
pages={34808 - 34820},
year={2023},
publisher={IEEE}
}

@article{allam2023,
title={A New Approach of Transparent and Explainable Artificial Intelligence Technique for Patient-Specific {ECG} Beat Classification},
author={Allam Jaya Prakash and others},
journal={IEEE Sensors Letters},
volume={7},
number={5},
year={2023},
publisher={IEEE}
}

@inproceedings{yang2023,
  title={A Dual-Scale Lead-Separated Transformer for {ECG} Classification},
  author={Yang Li and others},
  booktitle={Annual International Conference of the IEEE Engineering in Medicine \& Biology Society (EMBC)},
  year={2023},
  organization={IEEE}
}

@inproceedings{wei2024,
  title={A Multi-Resolution Mutual Learning Network for Multi-Label {ECG} Classification},
  author={Wei Huang and others},
  booktitle={International Conference on Bioinformatics and Biomedicine (BIBM)},
  year={2024},
  organization={IEEE}
}

@article{hany2024,
title={{ECGT}rans{F}orm: Empowering adaptive {ECG} arrhythmia classification framework with bidirectional transformer
},
author={Hany El-Ghaish and Emadeldeen Eldele},
journal={Biomedical Signal Processing and Control},
volume={89},
year={2024},
publisher={Elsevier}
}

@misc{tang2025,
      title={Hierarchical Transformer for Electrocardiogram Diagnosis}, 
      author={Xiaoya Tang and Jake Berquist and Benjamin A. Steinberg and Tolga Tasdizen},
      year={2025},
      eprint={2411.00755},
      archivePrefix={arXiv},
      primaryClass={cs.LG},
      url={https://arxiv.org/abs/2411.00755}, 
}

@inproceedings{xiaoyu2021,
  title={Ba{T}: Beat-aligned Transformer for Electrocardiogram Classification},
  author={Xiaoyu Li and others},
  booktitle={International Conference on Data Mining (ICDM)},
  year={2021},
  organization={IEEE}
}

@misc{tishby2000informationbottleneckmethod,
      title={The information bottleneck method}, 
      author={Naftali Tishby and Fernando C. Pereira and William Bialek},
      year={2000},
      eprint={physics/0004057},
      archivePrefix={arXiv},
      primaryClass={physics.data-an},
      url={https://arxiv.org/abs/physics/0004057}, 
}

@misc{tishby2015deeplearninginformationbottleneck,
      title={Deep Learning and the Information Bottleneck Principle}, 
      author={Naftali Tishby and Noga Zaslavsky},
      year={2015},
      eprint={1503.02406},
      archivePrefix={arXiv},
      primaryClass={cs.LG},
      url={https://arxiv.org/abs/1503.02406}, 
}

@inproceedings{almudévar2025aligningmultimodalrepresentationsinformation,
  title={Aligning Multimodal Representations through an Information Bottleneck},
  author={Antonio Almudévar and José Miguel Hernández-Lobato and Sameer Khurana and Ricard Marxer and Alfonso Ortega},
  booktitle={International Conference on Machine Learning (ICML)},
  year={2025}
}

@inproceedings{gopal20213kgcontrastivelearning12lead,
  title={{3KG}: Contrastive Learning of 12-Lead Electrocardiograms using Physiologically-Inspired Augmentations},
  author={Bryan Gopal and Ryan W. Han and Gautham Raghupathi and Andrew Y. Ng and Geoffrey H. Tison and Pranav Rajpurkar},
  booktitle={Advances in Neural Information Processing Systems (NeurIPS)},
  year={2021}
}

@inproceedings{clocs2021,
  title={{CLOCS}: Contrastive Learning of Cardiac Signals Across Space, Time, and Patients},
  author={Dani Kiyasseh and Tingting Zhu and David A. Clifton },
  booktitle={International Conference on Machine Learning (ICML)},
  year={2021}
}

@inproceedings{crystal2022,
  title={Contrastive Heartbeats: Contrastive Learning for Self-Supervised {ECG} Representation and Phenotyping},
  author={Crystal T. Wei and Ming-En Hsieh and Chien-Liang Liu and Vincent S. Tseng},
  booktitle={IEEE International Conference on Acoustics, Speech and Signal Processing (ICASSP)},
  year={2022}
}

@ARTICLE{pritam2022,
  author={Sarkar, Pritam and Etemad, Ali},
  journal={IEEE Transactions on Affective Computing}, 
  title={Self-Supervised ECG Representation Learning for Emotion Recognition}, 
  year={2022},
  volume={13},
  number={3},
  pages={1541-1554},
  doi={10.1109/TAFFC.2020.3014842}}

@inproceedings{Soltanieh2022,
  title={Analysis of Augmentations for Contrastive {ECG} Representation Learning},
  author={Sahar Soltanieh and Ali Etemad1 and Javad Hashem},
  booktitle={International Joint Conference on Neural Networks (IJCNN)},
  year={2022}
}

@ARTICLE{MaeFE2022,
  author={Zhang Huaicheng and others},
  journal={IEEE Transactions on Instrumentation and Measurement}, 
  title={Mae{FE}: Masked Autoencoders Family of Electrocardiogram for Self-Supervised Pretraining and Transfer Learning}, 
  year={2022},
  volume={72},
  number={},
  pages={1-15},
  doi={10.1109/TIM.2022.3228267}}

@ARTICLE{zhang2024,
  author={Zhang Wenrui and Yang Ling and Geng Shijia and Hong Shenda},
  journal={IEEE Transactions on Neural Networks and Learning Systems}, 
  title={Self-Supervised Time Series Representation Learning via Cross Reconstruction Transformer}, 
  year={2024},
  volume={35},
  number={11},
  pages={16129-16138},
  doi={10.1109/TNNLS.2023.3292066}}

@inproceedings{STMEM2024,
  title={Guiding Masked Representation Learning to Capture Spatio-Temporal Relationship of Electrocardiogram},
  author={Yeongyeon Na and Minje Park and Yunwon Tae and Sunghoon Joo},
  booktitle={International Conference on Learning Representations (ICLR)},
  year={2024}
}

@inproceedings{HeartLang2025,
  title={Reading Your Heart: Learning ECG Words and Sentences via Pre-training {ECG} Language Model},
  author={Jiarui Jin and others},
  booktitle={International Conference on Learning Representations (ICLR)},
  year={2025}
}

@misc{ecg-jepa-kim2024,
  title={Learning General Representation of 12-Lead Electrocardiogram with a Joint-Embedding Predictive Architecture},
  author={Sehun Kim},
  year={2024},
  eprint={2410.08559},
  archivePrefix={arXiv},
  primaryClass={cs.CV},
  url={https://arxiv.org/pdf/2410.08559}
}

@misc{ecg-jepa-kuba2024,
  title={Self-Supervised Pre-Training with Joint-Embedding Predictive Architecture Boosts {ECG} Classification Performance},
  author={Kuba Weimann and Tim O. F. Conrad},
  year={2024},
  eprint={2410.13867},
  archivePrefix={arXiv},
  primaryClass={cs.CV},
  url={https://arxiv.org/pdf/2410.13867}
}

@misc{ecg-fm2025,
  title={{ECG-FM}: An Open Electrocardiogram Foundation Model},
  author={Kaden McKeen and Sameer Masood and Augustin Toma and Barry Rubin and Bo Wang},
  year={2025},
  eprint={2408.05178},
  archivePrefix={arXiv},
  primaryClass={cs.CV},
  url={https://arxiv.org/pdf/2408.05178}
}

@misc{ecg-soup,
  title={{ECG-S}oup: Harnessing Multi-Layer Synergy for {ECG} Foundation Models},
  author={Phu X. Nguyen and others},
  year={2025},
  eprint={2509.00102},
  archivePrefix={arXiv},
  primaryClass={cs.CV},
  url={https://arxiv.org/pdf/2509.00102}
}

@inproceedings{junli2023,
  title={Frozen Language Model Helps {ECG} {Zero-Shot Learning}},
  author={Jun Li and Che Liu and Sibo Cheng and Rossella Arcucci and Shenda Hong},
  booktitle={Medical Imaging with Deep Learning (MIDL)},
  year={2023}
}

@inproceedings{merl2024,
  title={Zero-Shot {ECG} Classification with Multimodal Learning and Test-time Clinical Knowledge Enhancement},
  author={Che Liu and others},
  booktitle={International Conference on Machine Learning (ICML)},
  year={2024}
}

@article{
esi2024,
title={{ECG} Semantic Integrator ({ESI}): A Foundation {ECG} Model Pretrained with {LLM}-Enhanced Cardiological Text},
author={Han Yu and Peikun Guo and Akane Sano},
journal={Transactions on Machine Learning Research},
issn={2835-8856},
year={2024},
url={https://openreview.net/forum?id=giEbq8Khcf},
note={}
}

@inproceedings{dbeta2025,
  title={Boosting Masked {ECG}-Text Auto-Encoders as Discriminative Learners},
  author={Hung Manh Pham and Aaqib Saeed and Dong Ma},
  booktitle={International Conference on Machine Learning (ICML)},
  year={2025}
}

@inproceedings{melp2025,
  title={From Token to Rhythm: A Multi-Scale Approach for {ECG}-Language Pretraining},
  author={Fuying Wang and Jiacheng Xu and Lequan Yu},
  booktitle={International Conference on Machine Learning (ICML)},
  year={2025}
}

@misc{ecg-byte2024,
      title={{ECG-Byte}: A Tokenizer for End-to-End Generative Electrocardiogram Language Modeling}, 
      author={William Han and Chaojing Duan and Michael A. Rosenberg and Emerson Liu and Ding Zhao},
      year={2025},
      eprint={2412.14373},
      archivePrefix={arXiv},
      primaryClass={cs.CL},
      url={https://arxiv.org/abs/2412.14373}, 
}

@INPROCEEDINGS{ecg-chat2025,
  author={Zhao Yubao and Kang Jiaju and Zhang Tian and Han Puyu and Chen Tong},
  booktitle={IEEE International Conference on Multimedia and Expo (ICME)}, 
  title={{ECG-Chat}: A Large {ECG}-Language Model for Cardiac Disease Diagnosis}, 
  year={2025},
  volume={},
  number={},
  pages={1-6},
  doi={10.1109/ICME59968.2025.11209476}}

@inproceedings{qoq2025,
  title={{QoQ-M}ed: Building Multimodal Clinical Foundation Models with Domain-Aware {GRPO} Training},
  author={Wei Dai and Peilin Chen and Chanakya Ekbote and Paul Pu Liang},
  booktitle={Advances in Neural Information Processing Systems (NeurIPS)},
  year={2025}
}

@misc{lai2025medr1,
  title={{Med-R1}: Reinforcement Learning for Generalizable Medical Reasoning in Vision-Language Models},
  author={Yuxiang Lai and Jike Zhong and Ming Li and Shitian Zhao and Xiaofeng Yang},
  year={2025},
  eprint={2503.13939},
  archivePrefix={arXiv},
  primaryClass={cs.CV},
  url={https://arxiv.org/abs/2503.13939}
}

@inproceedings{hjelm2019learningdeeprepresentationsmutual,
  title={Learning deep representations by mutual information estimation and maximization},
  author={R Devon Hjelm and others},
  booktitle={International Conference on Learning Representations (ICLR)},
  year={2019}
}

@INPROCEEDINGS{lele2025,
  author={Chang Lele and Liu Peilin and Guo Qinghai and Wen Fei},
  booktitle={IEEE International Conference on Acoustics, Speech and Signal Processing (ICASSP)}, 
  title={Explicit Mutual Information Maximization for Self-Supervised Learning}, 
  year={2025},
  doi={10.1109/ICASSP49660.2025.10890783}}

@inproceedings{ting2020,
  title={A Simple Framework for Contrastive Learning of Visual Representations},
  author={Ting Chen and Simon Kornblith and Mohammad Norouzi and Geoffrey Hinton},
  booktitle={International Conference on Machine Learning (ICML)},
  year={2020},
  organization={PMLR}
}

@inproceedings{xiang2024,
  title={Towards Enhancing Time Series Contrastive Learning: A Dynamic Bad Pair Mining Approach},
  author={Xiang Lan and Hanshu Yan and Shenda Hong and Mengling Feng},
  booktitle={International Conference on Machine Learning (ICML)},
  year={2024},
  organization={PMLR}
}

@article{temesgen2022,
title={Self-supervised representation learning from 12-lead {ECG} data},
author={Temesgen Mehari and Nils Strodthoff},
journal={Computers in Biology and Medicine},
volume={141},
year={2022},
publisher={Elsevier}
}

@misc{oord2019representationlearningcontrastivepredictive,
      title={Representation Learning with Contrastive Predictive Coding}, 
      author={Aaron van den Oord and Yazhe Li and Oriol Vinyals},
      year={2019},
      eprint={1807.03748},
      archivePrefix={arXiv},
      primaryClass={cs.LG},
      url={https://arxiv.org/abs/1807.03748}, 
}

@article{Jin_2023,
   title={Med{CPT}: Contrastive Pre-trained Transformers with large-scale PubMed search logs for zero-shot biomedical information retrieval},
   volume={39},
   ISSN={1367-4811},
   url={http://dx.doi.org/10.1093/bioinformatics/btad651},
   DOI={10.1093/bioinformatics/btad651},
   number={11},
   journal={Bioinformatics},
   publisher={Oxford University Press (OUP)},
   author={Jin, Qiao and others},
   editor={Wren, Jonathan},
   year={2023},
   month=nov }

@article{PhysioNet-mimic-iv-ecg-1.0,
  author = {Gow, Brian and others},
  title = {{MIMIC-IV-ECG: Diagnostic Electrocardiogram Matched Subset}},
  journal = {{PhysioNet}},
  year = {2023},
  month = sep,
  note = {Version 1.0},
  doi = {10.13026/4nqg-sb35},
  url = {https://doi.org/10.13026/4nqg-sb35}
}

@article{PhysioNet-ptb-xl-1.0.3,
  author = {Wagner, Patrick and others},
  title = {{PTB-XL, a large publicly available electrocardiography dataset}},
  journal = {{PhysioNet}},
  year = {2022},
  month = nov,
  note = {Version 1.0.3},
  doi = {10.13026/kfzx-aw45},
  url = {https://doi.org/10.13026/kfzx-aw45}
}

@article{patrick2020,
title={{PTB-XL}, a large publicly available electrocardiography dataset},
author={Patrick Wagner and others},
journal={Scientific Data},
volume={7},
number={1},
page={154},
year={2020}
}

@article{Ng2018AnOA,
  title={An Open Access Database for Evaluating the Algorithms of Electrocardiogram Rhythm and Morphology Abnormality Detection},
  author={Eddie Y. K. Ng and Feifei Liu and Chengyu Liu and Lina Zhao and X. Zhang and Xiaoling Wu and Xiaoyan Xu and Yulin Liu and Caiyun Ma and Shoushui Wei and Zhiqiang He and Jianqing Li},
  journal={Journal of Medical Imaging and Health Informatics},
  year={2018},
  url={https://api.semanticscholar.org/CorpusID:70024401}
}

@article{PhysioNet-ecg-arrhythmia-1.0.0,
  author = {Zheng, Jianwei and Guo, Hangyuan and Chu, Huimin},
  title = {{A large scale 12-lead electrocardiogram database for arrhythmia study}},
  journal = {{PhysioNet}},
  year = {2022},
  month = aug,
  note = {Version 1.0.0},
  doi = {10.13026/wgex-er52},
  url = {https://doi.org/10.13026/wgex-er52}
}

@article{zheng2020,
title={Optimal Multi-Stage Arrhythmia Classification Approach},
author={Jianwei Zheng and others},
journal={Scientific Reports},
year={2020}
}

@inproceedings{10.5555/2981345.2981371,
author = {Barber, David and Agakov, Felix},
title = {The IM algorithm: a variational approach to Information Maximization},
year = {2003},
publisher = {MIT Press},
address = {Cambridge, MA, USA},
booktitle = {Proceedings of the 17th International Conference on Neural Information Processing Systems},
pages = {201–208},
numpages = {8},
location = {Whistler, British Columbia, Canada},
series = {NIPS'03}
}

\clearpage
\appendix
\section*{Appendix}

\section{Implementation Details}
\label{implementation_details}
\subsection{Pre-training Details}
\label{pretraining_details}
We use the MIMIC-ECG dataset \cite{PhysioNet-mimic-iv-ecg-1.0}, comprising 800{,}035 ECG-report pairs from 161{,}352 subjects. Each ECG recording is a 12-lead signal sampled at 500 Hz with a duration of 10 seconds. To reduce computational cost, all signals are downsampled to 100 Hz. We follow the preprocessing pipeline in~\cite{merl2024}, removing samples with invalid reports (e.g., empty or too short) and replacing abnormal values (NaN, Inf) with the average of neighboring points to ensure signal continuity.

\paragraph{Text Construction.}
Clinical reports are constructed by aggregating discrete diagnostic labels into a unified textual description. Specifically, individual diagnoses are concatenated into a single sentence (e.g., ``atrial fibrillation, left ventricular hypertrophy, ST depression''), providing a consistent semantic representation for multimodal learning.

\paragraph{Data Split.}
After preprocessing, the dataset contains 761{,}139 valid samples, split into 745{,}916 for training and 15{,}223 for validation.

\paragraph{Training Configuration. }
The model is trained with AdamW, an initial learning rate of $2 \times 10^{-4}$, and weight decay of $1 \times 10^{-5}$. Training is performed for 200 epochs with a cosine annealing learning rate schedule. The batch size is set to 128. All experiments are conducted on a single NVIDIA H100-80GB GPU.

\subsection{Downstream Details}
\label{downsteam_datasets}
\subsubsection{Downstream Datasets}
We evaluate on three public datasets: PTB-XL, CPSC2018, and the Chapman-Shaoxing-Ningbo (CSN) dataset. All signals are resampled to 100 Hz to match the pre-training setup.

\begin{itemize}
    \item \textbf{PTB-XL} contains 21{,}837 ECG recordings from 18{,}885 patients, with multi-label classification tasks including superclass, subclass, form, rhythm, and all-condition settings.
    \item \textbf{CPSC2018} contains 6{,}877 ECG recordings with 9 classification labels.
    \item \textbf{CSN} contains 45{,}152 recordings; after cleaning, 23{,}026 samples remain with 38 labels.
    \item \textbf{ECG Question-Answer (ECG-QA) dataset} was constructed to train the model to generate clinically meaningful textual interpretations directly from ECG waveforms. Each QA pair contains an ECG waveform, an instruction prompt, and a corresponding textual interpretation target. The target answers were generated using GPT-4o based on the original ECG diagnostic labels and reports, followed by expert review and refinement to ensure medical correctness and consistency. The generated responses describe rhythm characteristics, waveform abnormalities, conduction findings, and clinically relevant physiological observations in natural language form. During training, ECG waveforms were encoded by the ECG encoder and projected into the language model token space, while the language model was optimized using autoregressive next-token prediction on the target ECG interpretation text. The ECG-QA dataset follows the same split as the ECG dataset, consisting of 8,331 training QA pairs and 375 testing QA pairs.
\end{itemize}

\subsubsection{Downstream Evaluation}
\paragraph{Linear Probing.} 
In this setting, we utilize only the pre-trained ECG encoder, upon which a linear classification head is appended. All encoder parameters are fully frozen, and only the linear layer parameters are optimized during training. The model is trained using the AdamW optimizer with a fixed learning rate of 0.001 and a batch size of 64 for 100 epochs. Experiments are conducted on the aforementioned downstream datasets, including PTB-XL, which comprises five distinct classification tasks, as described earlier. To ensure reliability and reproducibility, we adopt a 10-fold cross-validation protocol. For PTB-XL, we follow the official 10-fold split. For the remaining datasets, we construct 10 folds with balanced label distributions. The final performance is reported as the average across all folds. This evaluation protocol assesses the quality of learned ECG representations, with a particular focus on their generalization and retention of clinically relevant physiological information across diverse tasks and datasets.

\paragraph{Zero-shot Evaluation.} 
In this setting, we leverage both ECG and text encoders, along with their corresponding projection heads, to perform zero-shot inference without additional training. Evaluation is conducted on the same downstream datasets and tasks as in the linear probing setup. However, since no model training is involved, we do not apply cross-validation and instead report performance on the entire dataset. Unlike linear probing, the dataset labels are not used directly as targets in multi-label classification; they are converted into clinical descriptions using the Clinical Knowledge Enhanced Prompt Engineering (CKEME) \cite{merl2024}, enabling evaluation in a shared semantic space. This setup allows us to assess the model’s ability to capture semantic alignment between ECG representations and textual concepts in a zero-shot setting.

\paragraph{Distribution-shift Evaluation.} We examine model robustness under distribution shift by constructing transfer settings across PTB-XL SuperClass, CPSC2018, and CSN, where each dataset is alternately treated as the source or target domain. A linear classifier is trained on top of the frozen ECG encoder using labeled data from the source domain, with model selection based on the existing train/validation splits in \cite{merl2024}. The training configuration, including the optimizer, learning rate, scheduler, and batch size, follows the same setup as the linear probing evaluation. The selected checkpoint is then directly evaluated on the full target dataset without any further adaptation. To ensure comparability across datasets, we align the label spaces between the source and target domains using a consistent category mapping and merging procedure, thereby preserving semantic equivalence across differing annotation schemes. This setup isolates the impact of distribution shift by assessing how well representations learned from one domain transfer to another with distinct data characteristics. The detailed category alignments used in cross-domain evaluation are summarized in Table~\ref{tab:domain_transfer_alignment}.

\begin{table}[H]
\centering
\scriptsize
\caption{Domain transfer category alignment across datasets.}
\label{tab:domain_transfer_alignment}
\small
\setlength{\tabcolsep}{16pt}
\renewcommand{\arraystretch}{0.95}

\resizebox{0.65\linewidth}{!}{
\begin{tabular}{cc}
\toprule
\textbf{PTB-XL Super} & \textbf{CPSC2018} \\
\midrule
HYP  & None \\
NORM & NORM \\
CD   & 1AVB, CRBBB, CLBBB \\
MI   & None \\
STTC & STE, STD \\

\midrule
\textbf{PTB-XL Super} & \textbf{CSN} \\
\midrule
HYP  & RVH, LVH \\
NORM & SR \\
CD   & 2AVB, 2AVB1, 1AVB, AVB, LBBB, RBBB, STDD \\
MI   & MI \\
STTC & STTC, STE, TWO, STTU, QTIE, TWC \\

\midrule
\textbf{CPSC2018} & \textbf{CSN} \\
\midrule
AFIB  & AFIB \\
VPC   & VPB \\
NORM  & SR \\
1AVB  & 1AVB \\
CRBBB & RBBB \\
STE   & STE \\
PAC   & APB \\
CLBBB & LBBB \\
STD   & STE, STTC, STTU, STDD \\
\bottomrule
\end{tabular}
}
\end{table}

\paragraph{Text Generation Evaluation.} To further evaluate the semantic quality and transferability of the learned ECG representations, we conduct an additional ECG-conditioned text generation experiment. Inspired by recent multimodal ECG-language modeling frameworks \cite{lai2025medr1}, we extend the pretrained ECG encoder with a LLM for autoregressive text generation.

Specifically, given an input ECG signal, the ECG encoder first extracts latent ECG representations, which are subsequently projected into the token embedding space of the LLM through a learnable ECG-to-LLM projection adapter. The projected ECG embeddings are then used as conditioning prefix tokens for autoregressive generation. We use the stage-1 generation setting from MedTVT-R1 as the evaluation protocol and adopt the same QA-style ECG-text generation task for comparison. The LLM backbone used in our experiments is LLaMA3.2-1B-Instruct.

During training, the ECG encoder parameters remain frozen, and only the ECG-to-LLM projection layers together with the LLaMA adapter parameters, including LoRA weights, biases, and normalization layers, are fine-tuned, following the Stage-1 training protocol of MedTVT-R1. We use the AdamW optimizer with an initial learning rate of $1\times10^{-4}$, cosine decay to $1\times10^{-6}$, and a weight decay of $0.05$. Each model is trained for 20 epochs without learning rate warmup using a batch size of 16. The maximum sequence length is set to 600 tokens.

To ensure fair comparison, all methods share the same LLM architecture, projection module, training configuration, and evaluation protocol. The only difference lies in the pretrained ECG encoder. We compare our proposed MERIT-pretrained encoder against the original MedTVT-R1 ECG encoder, as well as ECG encoders from QoQ \cite{qoq2025} and ECG-Chat \cite{ecg-chat2025}, each trained on the same LLM backbone under the same training setting.

The model is optimized using the standard autoregressive next-token prediction objective on ECG-question-answer pairs. Evaluation is conducted on the MedTVT-R1 QA test set using standard text generation metrics, including BLEU, METEOR, ROUGE, and BERTScore.

The displayed response and ground-truth texts in Figure~\ref{fig:text_generation_example} are shortened for readability. Below, we provide the original unprocessed texts corresponding to the same example.

\textbf{Original Ground Truth:}
\textit{In assessing the electrocardiogram, we observed several significant findings that provide insights into the patient’s cardiac health. Firstly, the presence of atrial fibrillation with rapid ventricular response suggests an irregular and often rapid heart rate originating from the atria. This condition can lead to symptoms such as palpitations, fatigue, or shortness of breath, and increases the risk of stroke and heart failure. Secondly, there is a possible left anterior fascicular block, also known as left anterior hemiblock. This refers to a delay or obstruction in the electrical conduction through the left anterior fascicle of the heart’s conduction system. It can result in a left axis deviation in the heart's electrical activity and may indicate underlying heart disease or be an incidental finding. Moreover, extensive ST-T changes are noted, which may be indicative of myocardial ischemia. This occurs when blood flow to the heart muscle is reduced, often due to coronary artery disease. The ST-T changes suggest that there may be areas of the heart that are not receiving adequate oxygen, which can lead to chest pain (angina) or even a heart attack if the blood supply is significantly compromised. In conclusion, the ECG reveals atrial fibrillation with a rapid ventricular response, which requires careful management to prevent complications. The potential left anterior fascicular block could signal an underlying conduction issue. The extensive ST-T changes raise concern for myocardial ischemia, necessitating further evaluation and possibly urgent intervention to prevent serious cardiac events. Overall, these findings contribute to an abnormal ECG, illustrating multiple areas of cardiac pathology that warrant comprehensive evaluation and management.}

\textbf{Original Response:}
\textit{In evaluating the provided ECG, several significant findings emerge. First, there is a first-degree atrioventricular block, suggesting a potential block in a pre-excitation syndrome, which requires attention. Firstly, there is evidence of atrial fibrillation with a rapid ventricular response. This condition is characterized by an irregular and often rapid heart rhythm originating from the atria. This can lead to symptoms such as palpitations, dizziness, or chest pain and may require further investigation if necessary. It’s essential to correlate this with the patient's symptoms symptoms and risk factors for cardiac health. Additionally, the ECG shows a left anterior fascicular block (LAFB), which indicates a delay or obstruction in the electrical conduction through the left anterior fascicle of the heart. This can occur in many individuals or associated with the heart's left, often associated with a left anterior fascicular block. LAFB is also noted, which may indicate an underlying cardiac condition that warrants further investigation and possibly intervention. Additionally, the lateral ST-T changes observed are observed. This can be attributed to the left ventricular hypertrophy or conduction abnormalities. The LBBB is often associated with a condition based on the observed anterior and lateral ST-T changes. These changes can be associated with ischemia or myocardial injury. It is crucial to correlate these ECG findings with the patient’s clinical presentation and history for a comprehensive assessment. In summary, the ECG demonstrates atrial fibrillation with rapid ventricular response, a leftward axis deviation of the heart's electrical axis and conduction. The combination of RBBB and LAFB is classified as abnormal, reflecting the collective abnormalities observed. These findings warrant further clinical correlation and potentially additional diagnostic testing to assess the underlying causes and guide management.}

\section{Extended Results}
\label{extended_eval}

\begin{table}[!htbp]
\centering
\scriptsize
\setlength{\tabcolsep}{1.5pt}
\renewcommand{\arraystretch}{0.82}
\begin{tabular}{lccccccccc}
\toprule
& \multicolumn{3}{c}{PTB-XL SuperClass} 
& \multicolumn{3}{c}{PTB-XL SubClass} 
& \multicolumn{3}{c}{PTB-XL Form} \\
\cmidrule(lr){2-4} \cmidrule(lr){5-7} \cmidrule(lr){8-10}
Methods 
& AUC & F1 & Accuracy
& AUC & F1 & Accuracy
& AUC & F1 & Accuracy \\
\midrule
ECG-FM \cite{ecg-fm2025}
& $87.58_{(0.01)}$ & $67.21_{(1.00)}$ & $84.10_{(0.56)}$
& $87.23_{(1.03)}$ & $41.68_{(1.99)}$ & $92.41_{(0.83)}$
& $78.19_{(2.48)}$ & $31.06_{(1.37)}$ & $85.26_{(2.60)}$ \\

STMEM \cite{STMEM2024} 
& $92.03_{(0.01)}$ & $73.52_{(0.97)}$ & $87.52_{(0.59)}$
& $91.74_{(0.01)}$ & $\cellcolor{gray!30} 49.82_{(1.78)}$ & $94.32_{(0.65)}$
& $\textbf{89.15}_\textbf{(0.02)}$ & $\textbf{45.68}_\textbf{(3.22)}$ & $\textbf{92.24}_\textbf{(1.44)}$ \\

MERL \cite{merl2024} 
& $91.20_{(0.01)}$ & $72.69_{(0.78)}$ & $87.19_{(0.60)}$
& $92.70_{(0.01)}$ & $49.54_{(1.54)}$ & $\cellcolor{gray!30} 94.95_{(0.40)}$
& $83.72_{(0.02)}$ & $36.47_{(2.17)}$ & $88.99_{(1.95)}$ \\

ESI \cite{esi2024}
& $85.98_{(0.01)}$ & $65.48_{(0.80)}$ & $82.98_{(0.81)}$
& $83.34_{(0.01)}$ & $33.68_{(0.75)}$ & $89.41_{(1.16)}$
& $77.34_{(0.03)}$ & $30.81_{(1.53)}$ & $84.19_{(3.38)}$ \\

D-BETA \cite{dbeta2025}
& $91.21_{(0.01)}$ & $72.56_{(1.04)}$ & $87.26_{(0.70)}$
& $91.66_{(0.01)}$ & $48.93_{(2.64)}$ & $94.94_{(0.60)}$
& $85.23_{(0.02)}$ & $39.16_{(1.54)}$ & $90.68_{(1.29)}$ \\

QoQ \cite{qoq2025}  
& $88.54_{(0.01)}$ & $68.80_{(0.62)}$ & $85.28_{(0.73)}$
& $87.96_{(0.01)}$ & $41.50_{(1.86)}$ & $93.42_{(0.65)}$
& $81.01_{(0.02)}$ & $31.78_{(2.32)}$ & $87.29_{(2.12)}$ \\

ECG-Chat \cite{ecg-chat2025}  
& $\cellcolor{gray!30} 92.25_{(0.00)}$ & $\cellcolor{gray!30} 74.40_{(0.49)}$ & $\cellcolor{gray!30} 87.96_{(0.57)}$
& $\cellcolor{gray!30} 91.74_{(0.01)}$ & $48.98_{(2.04)}$ & $94.94_{(0.63)}$ 
& $84.17_{(0.03)}$ & $39.58_{(1.45)}$ & $89.32_{(2.28)}$ \\
\midrule
Our 
& $\textbf{93.26}_{\textbf{(0.01)}}$ & $\textbf{75.97}_{\textbf{(0.93)}}$ & $\textbf{89.03}_\textbf{{(0.49)}}$
& $\textbf{93.79}_\textbf{{(0.01)}}$ & $\textbf{54.89}_\textbf{{(2.54)}}$ & $\textbf{95.68}_\textbf{{(0.48)}}$
& $\cellcolor{gray!30} 87.76_{(0.02)}$ & $\cellcolor{gray!30} 44.14_{(2.50)}$ & $\cellcolor{gray!30} 91.54_{(0.94)}$ \\
\bottomrule
\end{tabular}


\begin{tabular}{lcccccc}
\toprule
& \multicolumn{3}{c}{PTB-XL Rhythm}
& \multicolumn{3}{c}{PTB-XL All} \\
\cmidrule(lr){2-4} \cmidrule(lr){5-7}
Methods
& AUC & F1 & Accuracy
& AUC & F1 & Accuracy \\
\midrule
ECG-FM \cite{ecg-fm2025}
& $88.50_{(0.02)}$ & $54.41_{(3.51)}$ & $95.76_{(1.88)}$
& $83.87_{(0.01)}$ & $31.31_{(1.32)}$ & $92.26_{(0.85)}$  \\

STMEM \cite{STMEM2024}
& $\cellcolor{gray!30} 98.02_{(0.00)}$ & $\textbf{68.55}_\textbf{(3.44)}$ & $\cellcolor{gray!30} 98.61_{(0.11)}$
& $91.02_{(0.01)}$ & $39.30_{(1.09)}$ & $95.22_{(0.83)}$ \\

MERL \cite{merl2024} 
& $91.03_{(0.02)}$ & $53.27_{(2.75)}$ & $95.55_{(1.75)}$
& $88.97_{(0.01)}$ & $35.80_{(1.95)}$ & $93.88_{(1.10)}$ \\

ESI \cite{esi2024}
& $90.37_{(0.02)}$ & $39.69_{(1.90)}$ & $95.12_{(1.82)}$
& $74.81_{(0.01)}$ & $18.22_{(0.43)}$ & $84.19_{(1.45)}$ \\

D-BETA \cite{dbeta2025}
& $97.55_{(0.00)}$ & $65.77_{(3.81)}$ & $98.37_{(0.15)}$
& $\cellcolor{gray!30} 91.52_{(0.01)}$ & $\cellcolor{gray!30} 39.34_{(0.97)}$ & $\cellcolor{gray!30} 95.53_{(0.81)}$ \\

QoQ \cite{qoq2025}
& $88.41_{(0.02)}$ & $46.41_{(4.48)}$ & $94.95_{(2.14)}$
& $83.76_{(0.01)}$ & $28.55_{(1.62)}$ & $92.40_{(0.93)}$ \\

ECG-Chat \cite{ecg-chat2025}  
& $94.25_{(0.01)}$ & $58.76_{(3.47)}$ & $97.80_{(0.39)}$
& $88.01_{(0.01)}$ & $36.40_{(1.61)}$ & $94.42_{(1.09)}$ \\
\midrule
Our 
& $\textbf{98.03}_\textbf{(0.01)}$ & $\cellcolor{gray!30} 67.50_{(3.97)}$ & $\textbf{98.63}_\textbf{(0.22)}$
& $\textbf{92.70}_\textbf{(0.01)}$ & $\textbf{42.63}_\textbf{(1.08)}$ & $\textbf{95.80}_\textbf{(0.68)}$ \\
\bottomrule
\end{tabular}

\begin{tabular}{lcccccc}
\toprule
& \multicolumn{3}{c}{CSN}
& \multicolumn{3}{c}{CPSC} \\
\cmidrule(lr){2-4} \cmidrule(lr){5-7}
Methods
& AUC & F1 & Accuracy
& AUC & F1 & Accuracy \\
\midrule
ECG-FM \cite{ecg-fm2025}
& $88.70_{(0.01)}$ & $41.33_{(2.16)}$ & $96.26_{(0.46)}$
& $90.55_{(0.01)}$ & $67.18_{(1.49)}$ & $92.49_{(1.11)}$  \\

STMEM \cite{STMEM2024}
& $\cellcolor{gray!30} 95.35_{(0.01)}$ & $\cellcolor{gray!30} 52.83_{(1.25)}$ & $\cellcolor{gray!30} 97.52_{(0.17)}$
& $96.36_{(0.00)}$ & $79.96_{(0.74)}$ & $95.92_{(0.23)}$ \\

MERL \cite{merl2024}
& $91.32_{(0.01)}$ & $43.95_{(1.05)}$ & $95.79_{(0.39)}$ 
& $90.65_{(0.01)}$ & $66.84_{(1.20)}$ & $91.52_{(1.68)}$ \\

ESI \cite{esi2024}
& $77.01_{(0.02)}$ & $29.68_{(0.67)}$ & $85.28_{(2.93)}$ 
& $92.84_{(0.01)}$ & $70.99_{(0.91)}$ & $93.86_{(0.38)}$ \\

D-BETA \cite{dbeta2025}
& $95.16_{(0.00)}$ & $52.13_{(1.59)}$ & $97.49_{(0.17)}$ 
& $\textbf{96.51}_\textbf{(0.01)}$ & $\textbf{81.37}_\textbf{(1.25)}$ & $\textbf{96.20}_\textbf{(0.23)}$ \\

QoQ \cite{qoq2025}
& $85.41_{(0.01)}$ & $35.14_{(1.03)}$ & $93.04_{(1.13)}$
& $88.98_{(0.01)}$ & $63.23_{(1.55)}$ & $90.41_{(1.68)}$ \\

ECG-Chat \cite{ecg-chat2025}  
& $93.83_{(0.01)}$ & $50.42_{(2.29)}$ & $97.15_{(0.48)}$ 
& $95.06_{(0.00)}$ & $76.94_{(0.92)}$ & $95.28_{(0.30)}$ \\
\midrule
Our 
& $\textbf{96.46}_\textbf{(0.00)}$ & $\textbf{55.55}_\textbf{(1.44)}$ & $\textbf{97.82}_\textbf{(0.19)}$
& $\cellcolor{gray!30} 96.46_{(0.00)}$ & $\cellcolor{gray!30} 79.88_{(1.44)}$ & $\cellcolor{gray!30} 95.93_{(0.30)}$ \\
\bottomrule
\end{tabular}
\caption{Performance comparison on PTB-XL across multiple label taxonomies (Superclass, Sub-class, Form, Rhythm, and All conditions), CSN, and CPSC. Results are reported as mean $\pm$ standard deviation. The best results are highlighted in bold, and the second-best results are shaded in gray.}
\label{tab:ptbxl_results_detail}
\end{table}

\begin{table}[!htbp]
\centering
\scriptsize
\caption{Distribution-shift ECG classification results on AUC.}
\label{tab:cross_domain}
\begin{tabular}{l*{6}{c}}
\toprule
\multirow{2}{*}{\textbf{Source Domain}} \\
\multirow{2}{*}{\textbf{Target Domain}} 
& \multicolumn{2}{c}{\textbf{PTBXL-Super}}
& \multicolumn{2}{c}{\textbf{CPSC}}
& \multicolumn{2}{c}{\textbf{CSN}} \\
\cmidrule(lr){2-3} \cmidrule(lr){4-5} \cmidrule(lr){6-7}

& \textbf{CPSC} & \textbf{CSN}
& \textbf{PTBXL-Super} & \textbf{CSN}
& \textbf{PTBXL-Super} & \textbf{CPSC} \\
\midrule

ECG-FM~\cite{ecg-fm2025}
& 75.04 & 80.82 & \cellcolor{gray!30} 63.79 & 81.57 & 62.07 & 70.92 \\

STMEM~\cite{STMEM2024}
& 83.00 & 84.68 & 63.45 & 88.44 & 69.54 & 70.89 \\

MERL~\cite{merl2024}
& 85.46 & 87.44 & 61.33 & 83.65 & 69.92 & 71.37 \\

ESI~\cite{esi2024}
& 67.01 & 69.77 & 54.73 & 61.84 & 53.04 & 59.89 \\

D-BETA~\cite{dbeta2025}
& \textbf{86.61} & 87.55 & \textbf{66.54} & \cellcolor{gray!30} 90.26 & \cellcolor{gray!30} 71.91 & 75.55 \\

QoQ Encoder~\cite{qoq2025}
& 80.96 & 82.69 & 61.29 & 76.87 & 69.29 & 66.40 \\

ECG-Chat~\cite{ecg-chat2025}
& \cellcolor{gray!30} 85.87 & \textbf{89.82} & 60.31 & 88.60 & 67.18 & \textbf{78.82} \\

\midrule
\textbf{Ours}
& 85.83 & \cellcolor{gray!30} 88.52 & 61.88 & \textbf{90.36} & \textbf{73.44} & \cellcolor{gray!30} 76.51 \\

\bottomrule
\end{tabular}
\end{table}

\begin{table}[!htbp]
\centering
\scriptsize
\setlength{\tabcolsep}{1.5pt}
\renewcommand{\arraystretch}{0.82}
\caption{Ablation study under linear probing on PTB-XL, CSN, and CPSC.}
\label{tab:ablation_linear_detail}

\begin{tabular}{@{}lccc@{\hspace{0.2cm}}ccc@{\hspace{0.2cm}}ccc@{}}
\toprule
& \multicolumn{3}{c}{PTB-XL SuperClass} & \multicolumn{3}{c}{PTB-XL SubClass} & \multicolumn{3}{c}{PTB-XL Form} \\
\cmidrule(lr){2-4} \cmidrule(lr){5-7} \cmidrule(lr){8-10}
Methods & AUC & F1 & Accuracy & AUC & F1 & Accuracy & AUC & F1 & Accuracy \\
\midrule
MSE & $89.52_{(0.01)}$ & $69.58_{(0.85)}$ & $85.63_{(0.80)}$
    & $88.98_{(0.02)}$ & $44.32_{(1.64)}$ & $92.89_{(1.73)}$ 
    & $84.83_{(0.01)}$ & $37.23_{(1.64)}$ & $90.41_{(1.10)}$ \\
IB  & $92.59_{(0.01)}$ & $74.91_{(0.97)}$ & $88.56_{(0.57)}$
    & $92.92_{(0.01)}$ & $\cellcolor{gray!30} 52.77_{(1.72)}$ & $\cellcolor{gray!30} 95.20_{(0.76)}$ 
    & $\cellcolor{gray!30} 86.77_{(0.02)}$ & $40.66_{(1.59)}$ & $\cellcolor{gray!30} 90.63_{(1.79)}$ \\
CMA & $\cellcolor{gray!30} 92.67_{(0.01)}$ & $\cellcolor{gray!30} 75.17_{(0.89)}$ & $\cellcolor{gray!30} 88.69_{(0.58)}$
    & $\cellcolor{gray!30} 93.01_{(0.01)}$ & $52.46_{(2.16)}$ & $95.11_{(1.01)}$ 
    & $86.34_{(0.02)}$ & $\cellcolor{gray!30} 41.16_{(1.68)}$ & $89.87_{(2.26)}$ \\
Our & $\textbf{93.26}_\textbf{(0.01)}$ & $\textbf{75.97}_\textbf{(0.93)}$ & $\textbf{89.03}_\textbf{(0.49)}$
    & $\textbf{93.79}_\textbf{(0.01)}$ & $\textbf{54.89}_\textbf{(2.54)}$ & $\textbf{95.68}_\textbf{(0.48)}$
    & $\textbf{87.76}_\textbf{(0.02)}$ & $\textbf{44.14}_\textbf{(2.50)}$ & $\textbf{91.54}_\textbf{(0.94)}$ \\
\bottomrule
\end{tabular}


\begin{tabular}{@{}lccc@{\hspace{0.5cm}}ccc@{}}
\toprule
& \multicolumn{3}{c}{PTB-XL Rhythm} & \multicolumn{3}{c}{PTB-XL All} \\
\cmidrule(lr){2-4} \cmidrule(lr){5-7}
Methods & AUC & F1 & Accuracy & AUC & F1 & Accuracy \\
\midrule
MSE & $95.94_{(0.01)}$ & $60.93_{(3.61)}$ & $\cellcolor{gray!30} 98.16_{(0.28)}$
    & $88.25_{(0.01)}$ & $34.04_{(1.12)}$ & $94.11_{(0.48)}$ \\
IB  & $96.81_{(0.01)}$ & $\cellcolor{gray!30} 63.74_{(4.42)}$ & $98.11_{(0.32)}$
    & $91.67_{(0.01)}$ & $40.33_{(1.07)}$ & $95.23_{(0.83)}$ \\
CMA & $\cellcolor{gray!30} 96.89_{(0.01)}$ & $62.92_{(4.69)}$ & $98.09_{(0.25)}$
    & $\cellcolor{gray!30} 91.67_{(0.01)}$ & $\cellcolor{gray!30} 40.35_{(1.49)}$ & $\cellcolor{gray!30} 95.36_{(0.71)}$ \\
Our & $\textbf{98.03}_\textbf{(0.01)}$ & $\textbf{67.50}_\textbf{(3.97)}$ & $\textbf{98.63}_\textbf{(0.22)}$
    & $\textbf{92.70}_\textbf{(0.01)}$ & $\textbf{42.63}_\textbf{(1.08)}$ & $\textbf{95.80}_\textbf{(0.68)}$ \\
\bottomrule
\end{tabular}


\begin{tabular}{@{}lccc@{\hspace{0.5cm}}ccc@{}}
\toprule
& \multicolumn{3}{c}{CSN} & \multicolumn{3}{c}{CPSC} \\
\cmidrule(lr){2-4} \cmidrule(lr){5-7}
Methods & AUC & F1 & Accuracy & AUC & F1 & Accuracy \\
\midrule
MSE & $92.68_{(0.01)}$ & $42.97_{(0.87)}$ & $96.33_{(0.38)}$
    & $93.82_{(0.01)}$ & $73.03_{(1.39)}$ & $94.15_{(0.36)}$ \\
IB  & $95.39_{(0.01)}$ & $\cellcolor{gray!30} 51.44_{(1.15)}$ & $97.31_{(0.19)}$
    & $94.86_{(0.01)}$ & $75.30_{(1.35)}$ & $94.71_{(0.41)}$ \\
CMA & $\cellcolor{gray!30} 95.48_{(0.00)}$ & $51.33_{(1.62)}$ & $\cellcolor{gray!30} 97.40_{(0.26)}$
    & $\cellcolor{gray!30} 95.29_{(0.00)}$ & $\cellcolor{gray!30} 76.46_{(1.22)}$ & $\cellcolor{gray!30} 95.20_{(0.39)}$ \\
Our & $\textbf{96.46}_\textbf{(0.00)}$ & $\textbf{55.55}_\textbf{(1.44)}$ & $\textbf{97.82}_\textbf{(0.19)}$
    & $\textbf{96.46}_\textbf{(0.00)}$ & $\textbf{79.88}_\textbf{(1.44)}$ & $\textbf{95.93}_\textbf{(0.30)}$ \\
\bottomrule
\end{tabular}

\end{table}

\begin{table}[!htbp]
\centering
\scriptsize
\setlength{\tabcolsep}{1.5pt}
\renewcommand{\arraystretch}{0.82}
\caption{Ablation study under zero-shot evaluation on PTB-XL, CSN, and CPSC.}
\label{tab:ablation_zeroshot}
\renewcommand{\arraystretch}{1.1}

\begin{tabular}{lccccccccccccccc}
\toprule
& \multicolumn{3}{c}{PTB-XL SuperClass} 
& \multicolumn{3}{c}{PTB-XL SubClass} 
& \multicolumn{3}{c}{PTB-XL Form} 
& \multicolumn{3}{c}{PTB-XL Rhythm}
& \multicolumn{3}{c}{PTB-XL All} \\
\cmidrule(lr){2-4} \cmidrule(lr){5-7} \cmidrule(lr){8-10} \cmidrule(lr){11-13} \cmidrule(lr){14-16}
Methods 
& AUC & F1 & Accuracy 
& AUC & F1 & Accuracy 
& AUC & F1 & Accuracy 
& AUC & F1 & Accuracy
& AUC & F1 & Accuracy \\
\midrule

IB 
& \textbf{76.69} & \textbf{55.87} & 66.74
& \textbf{77.85} & \cellcolor{gray!30} 28.55 & \cellcolor{gray!30} 89.87
& \textbf{67.63} & 23.55 & \cellcolor{gray!30} 72.91 
& 80.80 & 30.99 & \cellcolor{gray!30} 90.44
& \cellcolor{gray!30} 74.52 & 19.24 & \textbf{87.82} \\

CMA 
& 76.21 & 55.03 & \textbf{76.39}
& 76.61 & 28.25 & 87.93
& 66.65 & \cellcolor{gray!30} 23.80 & \textbf{75.82} 
& \cellcolor{gray!30} 82.08 & \cellcolor{gray!30} 31.36 & 88.13 
& 74.2 & \cellcolor{gray!30} 19.47 & \cellcolor{gray!30} 87.21 \\

Our 
& \cellcolor{gray!30} 76.59 & \cellcolor{gray!30} 55.07 & \cellcolor{gray!30} 73.80
& \cellcolor{gray!30} 77.58 & \textbf{28.95} & \textbf{89.96}
& \cellcolor{gray!30} 66.89 & \textbf{24.17} & 64.68
& \textbf{82.76} & \textbf{32.96} & \textbf{90.48}
& \textbf{74.73} & \textbf{20.37} & 87.07 \\

\bottomrule
\end{tabular}


\begin{tabular}{lcccccc}
\toprule
& \multicolumn{3}{c}{CSN} 
& \multicolumn{3}{c}{CPSC} \\
\cmidrule(lr){2-4} \cmidrule(lr){5-7}
Methods 
& AUC & F1 & Accuracy 
& AUC & F1 & Accuracy \\
\midrule

IB 
& \cellcolor{gray!30} 78.43 & \cellcolor{gray!30} 23.97 & \cellcolor{gray!30} 87.20
& \textbf{85.03} & \textbf{54.87} & 86.35  \\

CMA 
& 77.76 & 23.85 & 85.13
& 84.97 & 54.45 & \textbf{87.67} \\

Our 
& \textbf{78.49} & \textbf{24.86} & \textbf{89.18}
& 83.20 & 52.18 & \cellcolor{gray!30} 86.45 \\

\bottomrule
\end{tabular}

\end{table}

\begin{figure}[!htbp]
    \centering

    \begin{subfigure}[t]{0.9\linewidth}
        \centering
        \includegraphics[width=\linewidth]{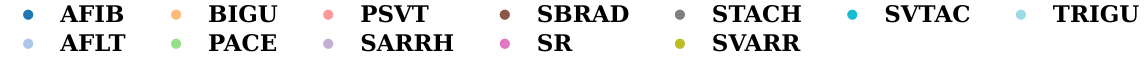}
    \end{subfigure}

    \vspace{0.5em}

    \begin{subfigure}[t]{0.24\linewidth}
        \centering
        \includegraphics[width=\linewidth]{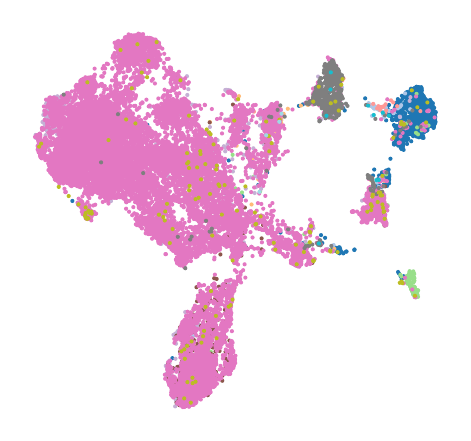}
        \caption{Our method.}
        \label{fig:sub1}
    \end{subfigure}
    \hfill
    \begin{subfigure}[t]{0.24\linewidth}
        \centering
        \includegraphics[width=\linewidth]{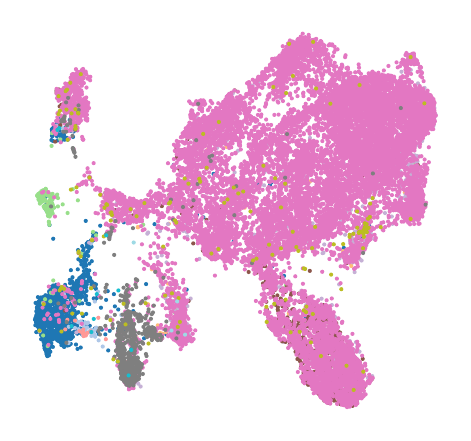}
        \caption{CMA.}
        \label{fig:sub2}
    \end{subfigure}
    \hfill
    \begin{subfigure}[t]{0.24\linewidth}
        \centering
        \includegraphics[width=\linewidth]{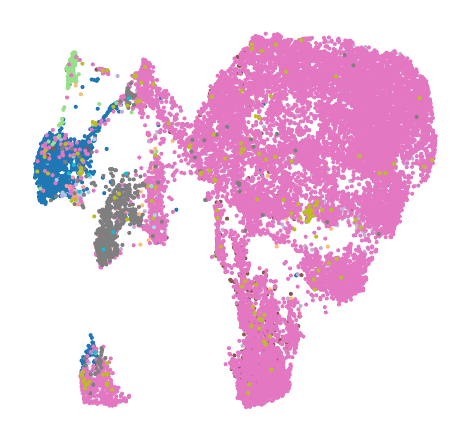}
        \caption{CMA + IB, $\lambda = 0.1$ \cite{almudévar2025aligningmultimodalrepresentationsinformation}.}
        \label{fig:sub3}
    \end{subfigure}
    \hfill
    \begin{subfigure}[t]{0.24\linewidth}
        \centering
        \includegraphics[width=\linewidth]{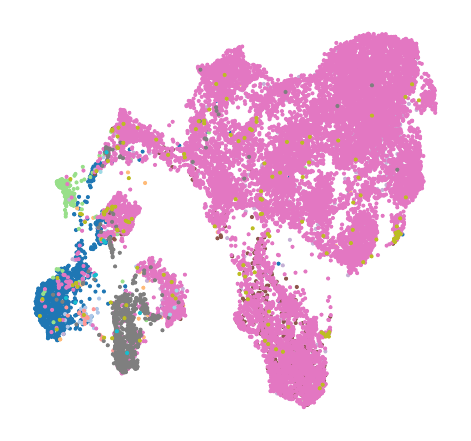}
        \caption{CMA + IB, $\lambda = 1$ \cite{almudévar2025aligningmultimodalrepresentationsinformation}.}
        \label{fig:sub4}
    \end{subfigure}

    \caption{\textbf{UMAP visualization of ECG representations on the PTB-XL Rhythm dataset}. ECG embeddings learned by different variants are projected into 2 dimensions and colored by the 12 rhythms. Compared with CMA and IB-based variants, the proposed MERIT framework produces more compact intra-class clusters and clearer inter-class separation across rhythm categories. For example, the rare rhythm PACE forms a more clearly distinguishable cluster under the proposed method. This suggests that our pretrained ECG representation captures rhythm-related physiological patterns more effectively, thereby improving discrimination between cardiac rhythm types.}
    \label{fig:umap}
\end{figure}

\section{Broader Impacts and Limitations}
\label{sec:impact}

ECG representation learning is directly relevant to clinical decision support, large-scale screening, and cardiology research. By learning representations that retain physiological structure while integrating clinical semantics, the proposed framework can improve diagnostic accuracy on fine-grained tasks and generalize better across hospitals, acquisition protocols, and patient cohorts, which is particularly valuable in settings where labeled data is scarce or annotation is expensive. The method may also reduce the cost of building specialized ECG models for downstream conditions and support clinician workflow through more reliable language-conditioned tools.

At the same time, models trained on a single pretraining corpus inherit the demographic and acquisition biases of that corpus: MIMIC-ECG is collected from a single U.S. hospital system and may underrepresent certain populations, age groups, or pathologies, which can translate into uneven downstream performance and, if deployed naively, exacerbate existing healthcare disparities. There is also a general risk of over-reliance on automated cardiac assessment without appropriate clinician oversight, and a risk of out-of-distribution failure when the model is applied to acquisition hardware, lead configurations, or sampling rates that differ from the training distribution. We therefore view the proposed framework as a research artifact intended for evaluation and as decision support, not as a stand-alone diagnostic system; any clinical deployment would require careful subgroup auditing, prospective validation, and human-in-the-loop oversight.

\clearpage

\end{document}